\documentclass[sigconf]{acmart}
\makeatletter
\renewcommand\@formatdoi[1]{\ignorespaces}
\makeatother
\renewcommand\footnotetextcopyrightpermission[1]{} 
\usepackage{setspace}
\usepackage{amsmath}
\usepackage{amsthm}
\usepackage{booktabs} 
\usepackage{multirow}
\usepackage{dcolumn}
\usepackage{algpseudocode}
\usepackage{algorithm}
\newfloat{algorithm}{t}{lop}
\usepackage{eqparbox}
\usepackage{threeparttable}
\usepackage{mathtools}
\usepackage{etoolbox}
\usepackage{siunitx}
\usepackage{subcaption}

\newcommand{\GNN} {{Con-GAE}}  
\newcount\Comments  
\usepackage{color}
\definecolor{darkgreen}{rgb}{0,0.5,0}
\definecolor{purple}{rgb}{1,0,1}
\newcommand{\kibitz}[2]{\ifnum\Comments=0\textcolor{#1}{#2}\fi}

\newcommand{\Yue}[1] {\kibitz{blue}     {[Yue: #1]}}

\AtBeginDocument{%
  \providecommand\BibTeX{{%
    \normalfont B\kern-0.5em{\scshape i\kern-0.25em b}\kern-0.8em\TeX}}}

\begin{document}

\title{Graph Convolutional Networks for traffic anomaly}

\author{Yue Hu}
\affiliation{%
  \institution{Vanderbilt University}
  \streetaddress{1025 16th Ave S}
  \city{Nashville}
  \state{Tennessee}
  \country{USA}
  \postcode{37212}
}
\email{yue.hu@vanderbilt.edu}

\author{Ao Qu}
\affiliation{%
  \institution{Vanderbilt University}
  \streetaddress{1025 16th Ave S}
  \city{Nashville}
  \state{Tennessee}
  \country{USA}}
\email{ao.qu@vanderbilt.edu}

\author{Dan Work}
\affiliation{%
  \institution{Vanderbilt University}
  \city{Nashville}
  \state{Tennessee}
  \country{USA}
}
\email{dan.work@vanderbilt.edu}

\renewcommand{\shortauthors}{Trovato and Tobin, et al.}

\begin{abstract}
Event detection has been an important task in transportation, whose task is to detect points in time when large events disrupts a large portion of the urban traffic network. Travel information {Origin-Destination} (OD) matrix data by map service vendors has large potential to give us insights to discover historic patterns and distinguish anomalies. However, to fully capture the spatial and temporal traffic patterns remains a challenge, yet serves a crucial role for effective anomaly detection. Meanwhile, existing anomaly detection methods have not well-addressed the extreme data sparsity and high-dimension challenges, which are common in OD matrix datasets. To tackle these challenges, we formulate the problem in a novel way, as detecting anomalies in a set of directed weighted graphs representing the traffic conditions at each time interval. We further propose \textit{Context augmented Graph Autoencoder} (\textbf{Con-GAE }), that leverages graph embedding and context embedding techniques to capture the spatial traffic network patterns while working around the data sparsity and high-dimensionality issue. Con-GAE adopts an autoencoder framework and detect anomalies via semi-supervised learning. Extensive experiments show that our method can achieve up can achieve a 0.1-0.4 improvements of the area under the curve (AUC) score over state-of-art anomaly detection baselines, when applied on several real-world large scale OD matrix datasets.
\end{abstract}

\maketitle
\thispagestyle{empty}
\pagestyle{plain}

\section{Introduction}
Event detection has been a long standing task in transportation~\cite{liu2011discovering,yang2014detecting}. Usually, when large events take place, a large portion of the urban traffic network is disrupted, resulting in anomalous traffic conditions.  Accurate and timely detection of such events can help with real-time resource allocation and congestion mitigation, enabling informed urban traffic management. Therefore, the objective of our work is to detect extreme traffic events, Which are anomalies that manifest on large portions of the network at once.

Today, abundant transportation data facilitates real time traffic monitoring and anomaly detection, with the aid of machine learning techniques. One promising area is in urban mobility, where large fleets of instrumented vehicles (e.g., ride sharing services and taxis) collect abundant traffic information, which is then processed into \textit{origin-destination} (OD) travel time data and made publicly available to assist urban mobility managers. Examples include data products from Uber Movement~\cite{UM}, taxi datasets in Chicago~\cite{taxi} and New York City~\cite{NYC}, and micro-mobility data from bike share and scooter operators (e.g., \cite{scooter}).

Despite the growing availability of such OD data, it is often challenging for urban mobility managers to directly use the data for performance monitoring and event detection. This is because the data is highly structured in time and space and also high dimensional. It is also sparse with many origin-destination pairs lacking data at any given moment in time.  The spatio-temporal structure, high dimensionality, and sparsity of the data inhibit standard anomaly detection approaches from achieving performance necessary for widespread adoption in current traffic management centers.



To address these challenges, we propose a \textit{Context augmented graph autoencoder} (\textbf{Con-GAE}) that leverages graph embedding and context embedding techniques to capture the spatio-temporal traffic network patterns. The autoencoder uses graph convolutional layers modified to account for asymmetry in travel times to capture spatial patterns in the data, even when many of the travel time entries are missing. It also uses context embedding to capture daily and weekly periodicity in the data, to further enhance performance. Finally, Con-GAE adopts a hierarchical structure to aggregate the spatial and temporal embeddings of the traffic network at each time step into a single comprehensive graph embedding, to  detect time contextualized anomaly graphs as a whole.  Through comparisons with existing state of the art anomaly detection methods and through an ablation study, we show that Con-GAE can achieve performance improvements to the \textit{area under the curve} (AUC) of 0.1 to 0.4 when detecting large scale anomalies in OD travel time data. 

\Yue{Explanation tackling missing and high-dimensionality?}

We give a high-level overview of the  structure that enables~\GNN~to fully capture the spatial correlations between city zones. Intuitively, Zone A and Zone B can have similar travel times to other parts of the network if the zones are physically close together, or close in travel time. We consider physical closeness by using geographic information as input node features, and consider the travel-time connection via edge weights. Then \GNN~learns the vector representation of each node given both node and weight information, such that Zone A and Zone B have similar embeddings when they are physically close or close in travel time. The node embeddings are further used to decode the edge weights, so two origin zones with similar embeddings will lead to similar travel times to other nodes in the network.

Our work is inspired by two threads of recent development in deep learning, namely graph convolutional networks and autoencoder-based anomaly detectors. The graph convolutional layers in \GNN~are a modification of the \textit{graph convolutional network} (GCN)~\cite{kipf2017semi}. Given a graph, GCN learns an embedding of the nodes in the graph that encodes structural information about the graph via aggregation of the embeddings of neighboring nodes on the graph.
While modified GCN layers can capture the geometric structure of the network, the dynamic nature of traffic requires further consideration of the temporal dimension. Existing works on dynamic graphs~\cite{zhou2018dynamic,goyal2018dyngem} mainly focus on modeling the short-term dependencies of consecutive graph instances. In comparison, we focus on exploiting the temporal periodicity of traffic graphs over longer time horizons. Daily or weekly periodicity in the data can better support traffic anomaly detection than short term fluctuations.  We further emphasize that while some approaches detect anomalous nodes and edges in dynamic graph~\cite{yu2018netwalk, zheng2019addgraph}, our task of extreme event detection requires detecting anomalous graph as a whole, rather than a part of a graph as anomaly.  


Con-GAE determines anomalies via an \textit{autoencoder} (AE) framework~\cite{sakurada2014anomaly} for semi-supervised anomaly detection. An AE consists of an encoder to compress the high-dimensional input into a low dimensional space, and a decoder to map the low dimensional representation back to the original input space minimizing the reconstruction error. When trained only on normal data, AE is assumed to encode and decode normal data well, but not anomalous data. Thus during testing, samples with large reconstruction errors are marked as anomalies.
AE-based anomaly detectors have shown success in various high-dimensional anomaly detection tasks \cite{zhou2017anomaly,chong2017abnormal, xu2018unsupervised}. Here we specialize the AE network architecture to exploit spatio-temporal structure in mobility data.

We show the effectiveness of our model on large-scale real-world dataset. We apply Con-GAE to an Uber movement OD travel time dataset, to detect two kinds of synthesized anomalies, i.e., large spatial anomalies impacting the network, and traffic conditions which are otherwise normal but not for the given time. Such temporal anomalies occur, e.g., due to holidays. We vary the magnitude of the anomalies and the number of anomalous instances, and show that our model outperforms several state-of-the-art baseline methods for high-dimensional or spatio-temporal anomaly detection. Finally, we show how the model reveals the influence of a real large scale event contained in the dataset.

In summary, our contributions are as follows:
\begin{itemize}
    \item We formulate the problem of anomaly detection in mobility OD datasets in a novel way, as detecting anomalies in a set of time dependent directed weighted graphs. Each graph has geographic information contained in the node features and travel time information in the edge weights.
    \item We propose Con-GAE, which is an autoencoder-based anomaly detector. Con-GAE uses GCN layers to generate node embeddings that capture spatial relationships, and time embeddings to capture the temporal periodicity. 
    \item Experiments on an OD mobility dataset demonstrating Con-GAE outperforms several state-of-the-art anomaly detection methods.
\end{itemize}


\section{Methodology}
In this section, we first model the traffic network as a graph and formulate the anomaly detection problem. Then we address the graph encoder and decoder structures. Finally we address the loss function and regularization strategies used.

\subsection{Modeling mobility data as a graph}
\begin{figure}
\centering



\begin{subfigure}[t]{0.46\columnwidth}
    \includegraphics[width=\linewidth]{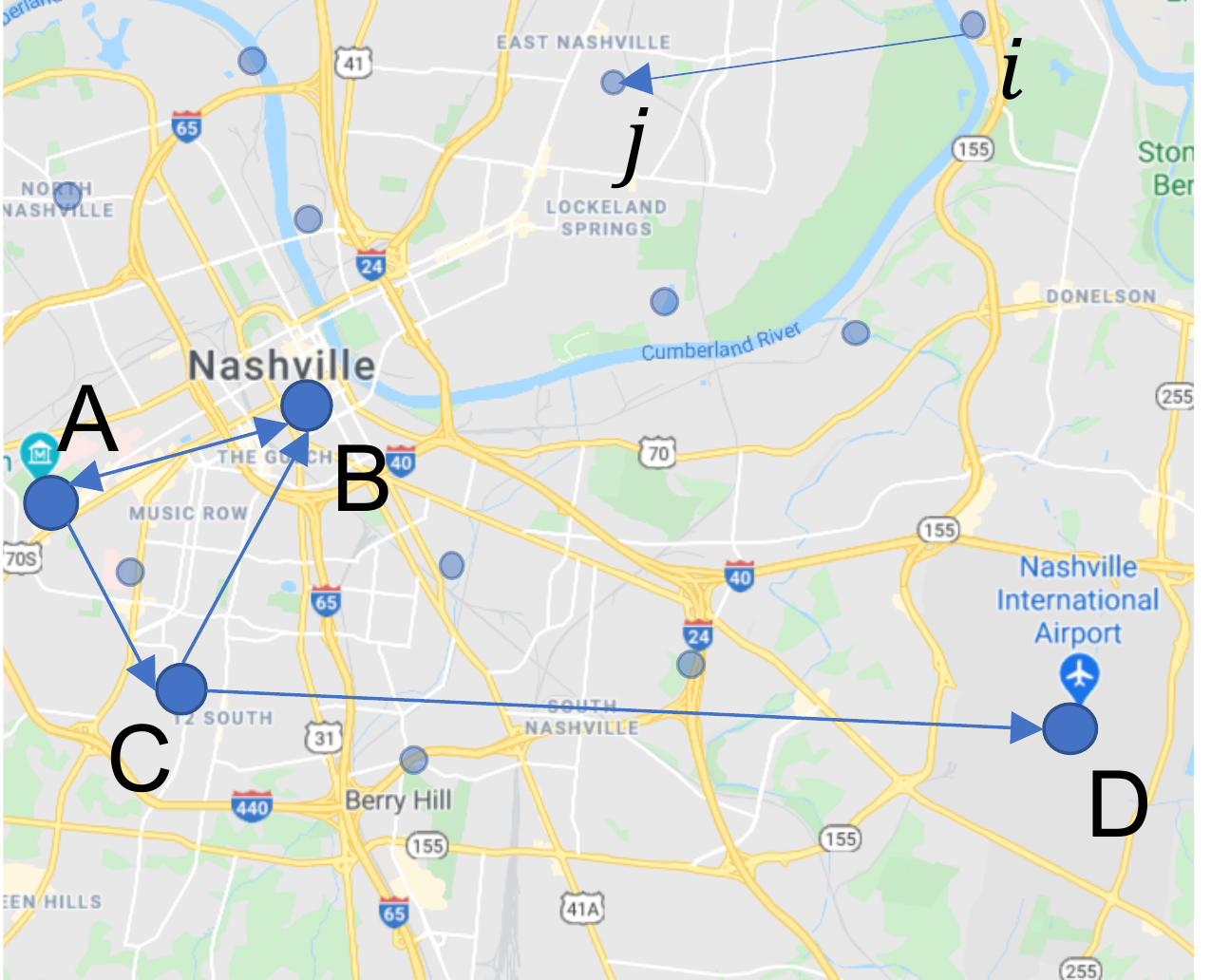}
    \caption{Traffic network graph}
    \label{fig:traiffic_graph}
\end{subfigure}
\quad
\begin{subfigure}[t]{0.48\columnwidth}
    \includegraphics[width=\linewidth]{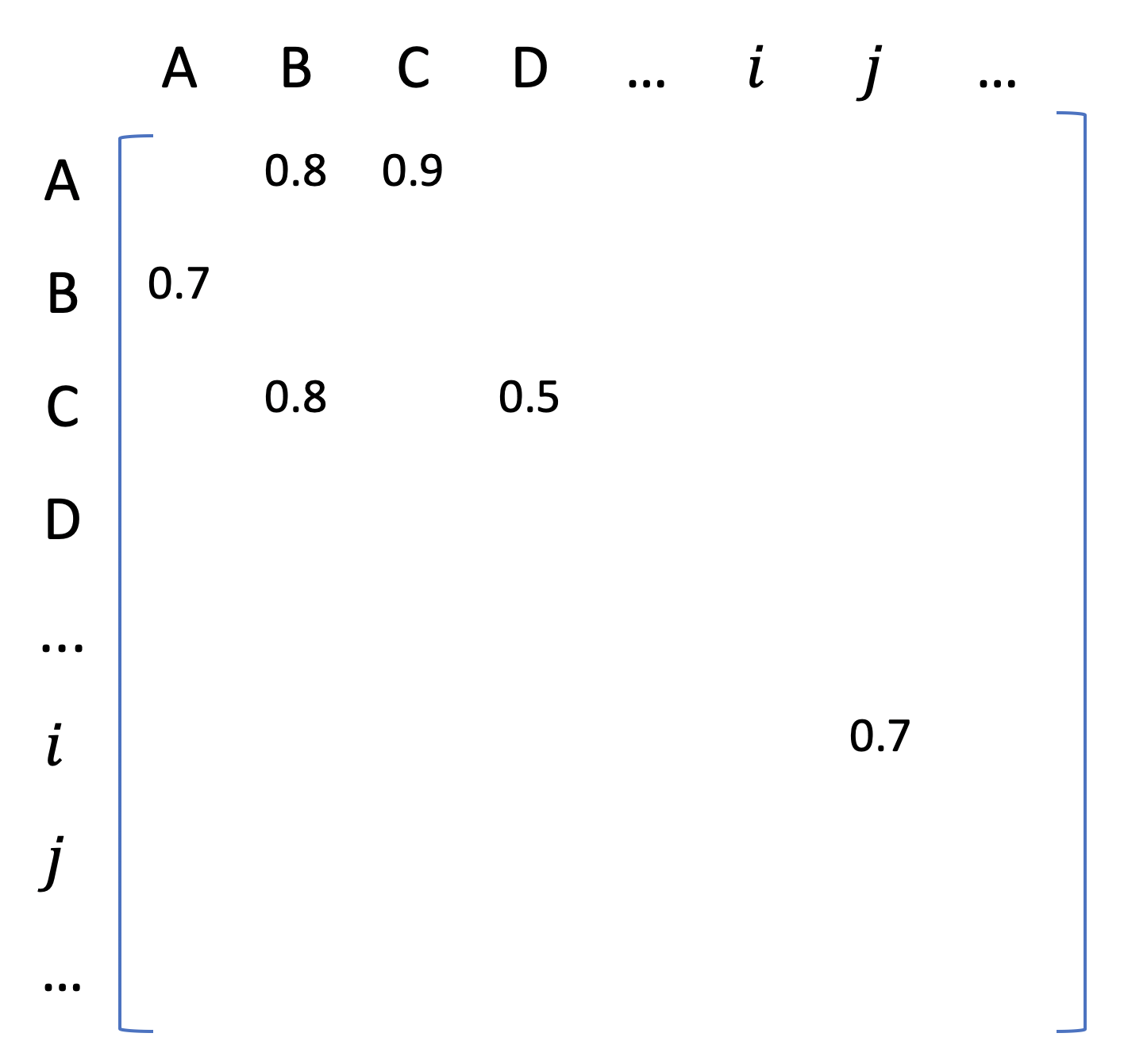}
    \caption{Weighted adjacency matrix}
    \label{fig:adj}
\end{subfigure}
\caption{Illustration of OD mobility graph (left) and corresponding weighted adjacency matrix (right) corresponding to scaled inverse travel times between points on the graph. The weighted adjacency matrix can be highly sparse due to lack of data on all OD pairs.}
\label{fig:intro}
\end{figure}

We introduce the graph model of the OD mobility data, and then formulate the anomaly detection problem on the graph. The mobility data can be modeled as a set of time dependent directed weighted graphs. At discrete time $t\in \left\{1,2,...,T\right\}=\mathcal{T}$, the graph is denoted as $ G(t) = (\mathcal{V}, \mathcal{E}(t),\mathbf{W}(t))$. The node set $\mathcal{V}$ is time invariant, representing the $N = |\mathcal{V}|$ distinct geographical zones in the city. There is an edge $e_{ij} \in \mathcal{E}(t)$ if the travel time information is available from node $v_i \in \mathcal{V}$ to  $v_j \in \mathcal{V}$ at time $t$. The weighted adjacency matrix is denoted $W(t)\in\mathbb{R}^{N \times N}$, where each element $w_{ij}(t)$ contains the scaled inverse travel time from node $v_i$ to $v_j$ at time $t$, such that larger values indicate shorter travel times. The geographic information for each zone is summarized in the node feature matrix $\mathbb{X} \in \mathbb{R}^{N \times d}$, which could include, e.g., location, size, or land use.  An OD  graph is shown in Fig~\ref{fig:intro}(a), with the corresponding adjacency matrix $\mathbf{W}(t)$ illustrated in Fig~\ref{fig:intro}(b). 

Our goal is the following. Given a set of graphs $\left\{G(t)|t\in\mathcal{T}\right\}$, find the anomalous times $t_a\in\mathcal{T}$ corresponding to large scale events disrupting the transportation network. We do so by training a context-augmented graph autoencoder that encodes each graph as a low-dimensional embedding conditioned on the time information, and decodes the embedding so as to minimize the average reconstruction error of the weighted adjacency matrix $\mathbf{W}(t)$. The autoencoder is trained only on normal data, so it  produces high reconstruction errors when events occur.


\subsection{Encoder}

\begin{figure}
    \centering
    \includegraphics[width = \columnwidth]{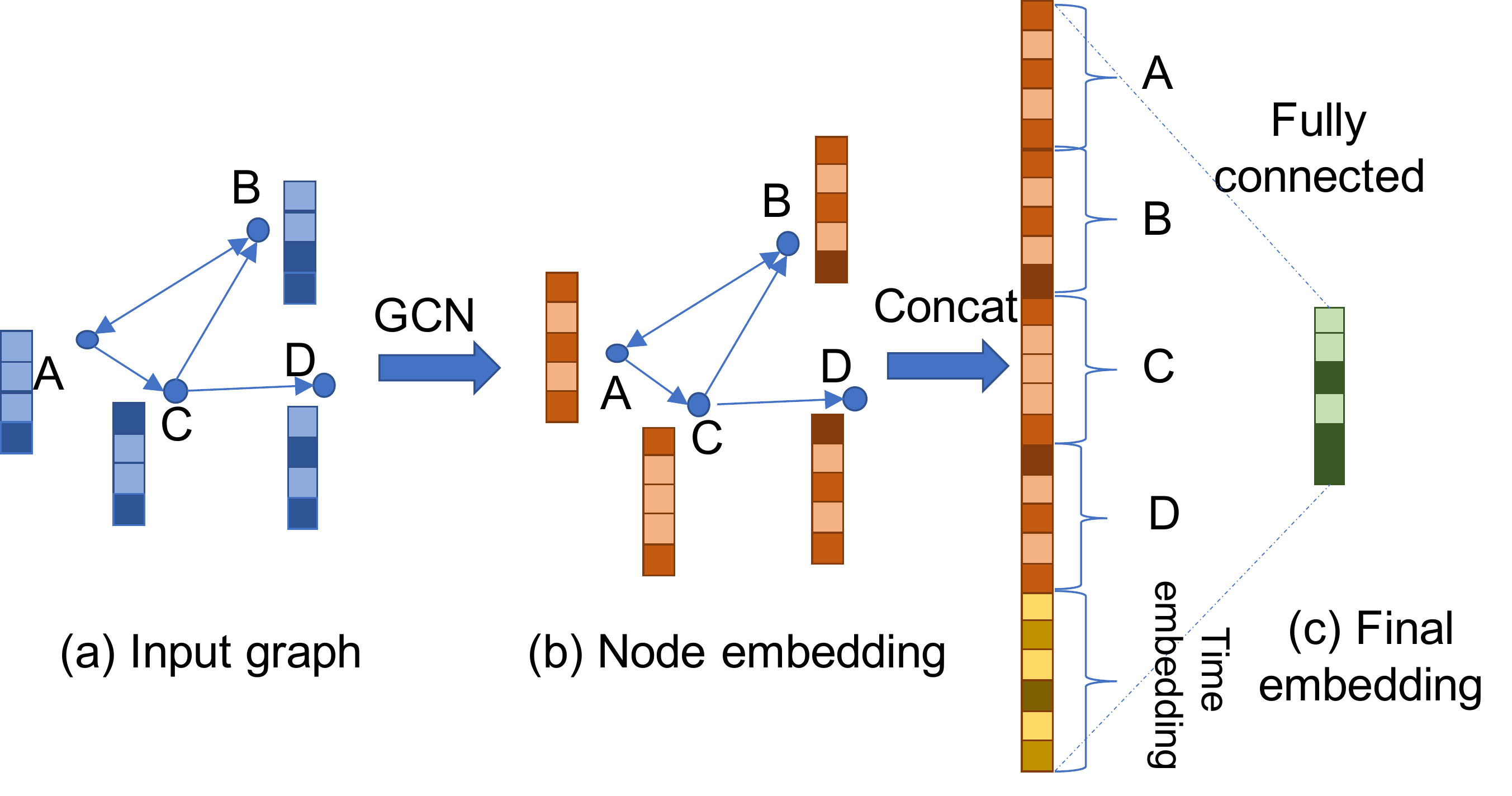}
    \caption{Illustration of encoder. We first apply GCN layers to aggregate the node features and edge weight information and calculate node embedding (b). Then we concatenate the node embedding and time embedding, and further shrink the embedding dimension, to calculate the embedding of the graph as a whole as in (c).}
    \label{fig:encoder}
\end{figure}

In an overview, the encoding process consists of three steps, as in Fig~\ref{fig:encoder}. We first apply GCN layers to encode the graph structure and node feature information, and learn an embedding vector for each node in the graph. We also use a distinct time embedding for each hour in the day and each day in the week to capture temporal information in the network. The node embeddings and the time embeddings are stacked in a vector, which is passed through a fully connected layer to arrive at a graph embedding of reduced dimension.

In the first step, we use a modified GraphSAGE~\cite{hamilton2017inductive} implementation of GCN to learn the node embedding, by aggregating the adjacent node information in a layer-by layer manner. The number of GraphSAGE layers stacked decides how many hops of neighborhood information is passed to each node~\cite{dwivedi2020benchmarking}. Let $\mathcal{N}_i(t)$ denote the neighbors of node $i$. GraphSAGE propagates the node $v_i$ embedding $h_i^l(t) \in \mathbb{R}^{d_l}$ in layer $l$ of the network to the embedding a $h_i^{l+1}(t) \in \mathbb{R}^{d_{l+1}}$ as:
\begin{equation}
\begin{aligned}
\label{eq:SAGE}
    \hat{h}_i^{l+1}(t) &= \text{ReLU}\left(U^l \text{Concat}(h_i^l(t), \text{Agg}_{v_j \in \mathcal{N}_i(t)}h_j^l(t))\right),\\
\end{aligned}
\end{equation}
where $\hat{h}_i^{l+1}(t)$ is the non-normalized node embedding, Concat is the concatenation operator, and Agg is an aggregation operator described subsequently. The normalized node $v_i$ embedding at layer $l+1$ is obtained by projecting onto the unit ball:         $h_i^{l+1}(t) = \hat{h}_i^{l+1}(t) / \| \hat{h}_i^{l+1}(t) \|_2$.

The standard GraphSAGE~\cite{hamilton2017inductive} aggregation operators for binary undirected edges aggregate information from the embeddings of the neighbors of $v_i$ via mean, max-pooling, or LSTM operations.  To account for the time varying weighted directed edges, we apply a weighted mean aggregation function. We take $\mathcal{N}_i(t)$ as the set of nodes $j$ for which a weighted edge $e_{ji}$ is present at time $t$. Then the weighted mean aggregation operator reads:

\begin{equation}
\begin{aligned}
\label{eq:dirSAGE}
    \text{Agg}_{v_j \in \mathcal{N}_i(t)}h_j^l(t):=\frac{ \sum_{j\in\mathcal{N}_j(t)} w_{ji}(t) h_j^l(t)}{\sum_{j\in\mathcal{N}_j(t)} w_{ji}(t)}.
\end{aligned}
\end{equation}

In the second step, we introduce context embeddings that that capture the weekly and hourly periodicity of the mobility data. We define context as additional information not contained in the OD matrix that the encoder and decoder is conditioned on. In our case we use temporal information as context. Context is implemented by concatenating two fixed length vectors: one is the embedding of the day of week information, the other is the embedding of the hour of the day. The embeddings capture our prior that traffic conditions on a specific time of day or a specific day of the week should be similar. The hour embedding is denoted as $h_{\text{hour}}\in \mathbb{R}^{d_\text{hour}}$, with one embedding per hour of day; similarly the week embedding is denoted as $h_{\text{week}} \in \mathbb{R}^{d_\text{week}}$, with one embedding per day of the week. Given the graph at time $t$ (e.g., 10am Tuesday Sept. 1), we use the corresponding $h_{\text{hour}}$ (i.e., 10 am) and $h_{\text{week}}$ (i.e., Tuesday) embedding. The entries of each embedding vector are initiated from i.i.d. normal distribution, then learned during training time.

In the last step, we combine the node and time embeddings into a single low-dimensional embedding through a fully connected layer:
\begin{equation}
\begin{aligned}
\label{eq:graph_emb}
    \widetilde{h}_G(t) &= \text{Concat}(h_1^L(t), h_2^L(t), \dots, h_N^L(t) ,h_{\text{hour}}(t), h_{\text{week}}(t)),\\
    h_G(t) &=  \text{ReLU}(U_G \widetilde{h}_G(t)),
\end{aligned}
\end{equation}
where $h_G(t) \in \mathbb{R}^{d_g}$ is the final graph embedding at time $t$, and $U_G \in \mathbb{R}^{d_g \times (Nd_L+d_\text{week}+d_\text{hour})}$ is weight matrix.


\subsection{Decoder}
\begin{figure}
    \centering
    \includegraphics[width = \columnwidth]{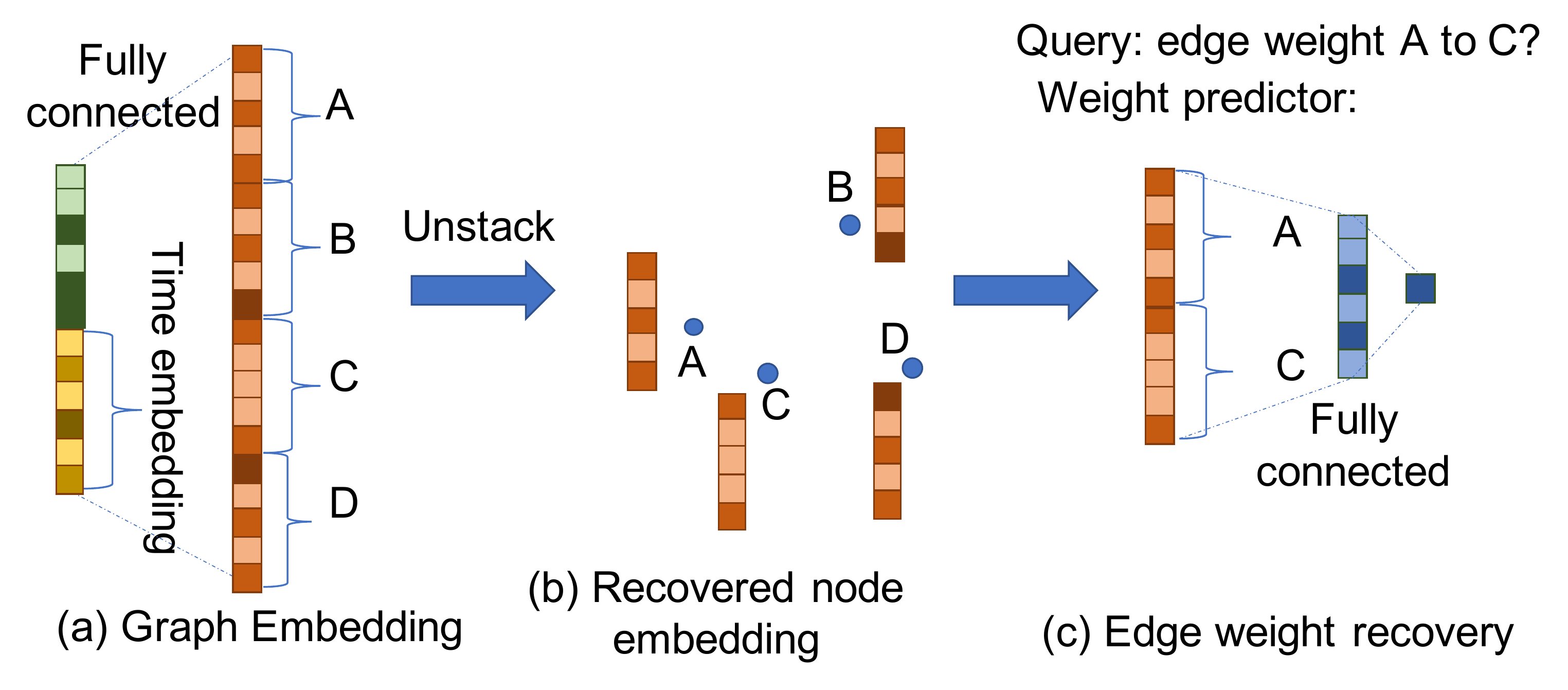}
    \caption{Illustration of decoder. We first calculate the node embeddings from graph encoding conditioned on time as in (a)(b). Then for each query of edge weight from one node to another, we  use a weight predictor to calculate from the two corresponding node embeddings (c).}
    \label{fig:decode}
\end{figure}

The decoder works in the opposite way as the encoder, as shown in Fig~\ref{fig:decode}.  We first calculate the node embedding from the graph embedding. Then, given each pair of node embeddings corresponding $v_i$ and $v_j$ at time $t$, we calculate the $w_{ij}(t)$ entry in the weighted adjacency matrix. 

We explicitly condition the graph decoding on time by concatenating the graph embedding $h_\mathcal{G}(t)$, with the corresponding time embeddings. Then a fully connected layer is used to recover a vector containing all node embeddings: 
\begin{equation}
\begin{aligned}
\label{eq:unstack}
    \widetilde{h}_G'(t) = \text{ReLU}(U_G' \text{Concat}(h_G(t),h_{\text{hour}}(t), h_{\text{week}}(t))), \\
    \left\{h_1(t), h_2(t), \dots, h_N(t)\right\} = \text{Unstack}(\widetilde{h}_G'(t)),
\end{aligned}
\end{equation}
which is subsequently unstacked to recover the individual node embeddings  $ h_i(t) \in \mathbb{R} ^{d_L} $. $U_G' \in \mathbb{R} ^{Nd_L \times (d+d_\text{week}+d_\text{hour})}$ is a weight matrix to be trained.

Given a pair of nodes $(v_i,v_j)$ and the corresponding recovered node embeddings $(h_i(t),h_j(t))$, we use an edge weight predictor to recover the edge weight $w_{ij}(t)$. Note that the predominantly used GAE edge predictor $w_{ij} =\sigma(h_i^Th_j)$ \cite{kipf2016variational} cannot be adopted here, because it produces symmetric results for $w_{ij}$ and $w_{ji}$. Because the graph corresponding to the mobility data is directed, forcing $w_{ij}, w_{ji}$ to be the same would introduce bias. A bilinear predictor $w_{ij} = h_i^TQh_j$ with $Q$ to be learned is also popular for asymmetric decoding~\cite{yang2014embedding}, but is slightly slower and performs slightly worse in our experiments.  Instead, we use  a parametrized MLP as a weight predictor, by first concatenating the two node embeddings and then applying fully connected layers:
\begin{equation}
\begin{aligned}
\label{eq:decode}
    \widetilde{w}_{ij}(t) &= \text{ReLU}(U^1_\text{dec} \text{Concat}(h_i(t),h_j(t))), \\
    w_{ij}'(t) &= \sigma(U^2_\text{dec} \widetilde{w}_{ij}(t)),
\end{aligned}
\end{equation}
where  $U^1_\text{dec} \in \mathbb{R} ^{d_e \times 2d_L}$ and $U^2_\text{dec} \in \mathbb{R} ^{1 \times d_e}$ are weight matrices. The final sigmoid layer $\sigma$ ensures $w'_{ij}(t)\in[0, 1]$. 

\subsection{Loss function}
We use the \textit{mean squared error} (MSE) between the entries of the original and recovered weighted adjacency matrices as the loss function. For graph instance $G(t)$  at time $t$, the loss function reads:
\begin{equation}
    \label{eq:MSE}
    \mathcal{L}(t) = \frac{1}{|\mathcal{E}(t)|}\sum_{e_{ij} \in \mathcal{E}(t)}(w_{ij}(t) - w'_{ij}(t))^2.
\end{equation}
We use the Adam optimizer~\cite{kingma2014adam} to minimize the loss function~\eqref{eq:MSE} for training instances in mini-batches. During training, the values of $U^1,\cdots,U^L$, $U_G,U'_G,U^1_\text{dec},U^2_\text{dec}$, the 24 $h_\text{hour}$, and the 7 $h_\text{week}$ time embeddings are updated. 
During testing, the loss function~\eqref{eq:MSE} for each testing instance is used as its anomaly score. 

\subsection{Regularization}
For regularization during training, we conduct edge dropout before feeding the graph data into Con-GAE, by randomly dropping edges from the edge set $\mathcal{E}(t)$ with probability $p_\text{e\_drop}$. Edge dropout amounts to masking a portion of edges and letting Con-GAE to predict edge weights without seeing the values. This can help improve the generalization ability, and force Con-GAE to learn the relations between nodes for correct prediction. We also apply dropout~\cite{srivastava2014dropout} with probability $p_\text{drop}$ on the output of the GraphSAGE layers as well as on the time embeddings. Dropout prevents over-reliance on  specific features that lead to locally optimal reconstruction solutions, such as the time embedding.

\section{Experiments}
We use a large-scale real-world transportation dataset to evaluate~\GNN. The experiments include: 1) a performance comparison of Con-GAE with other methods for spatial and temporal anomaly detection; 2) an ablation study to check the contribution of each component of Con-GAE; 3) a sensitivity study of the model hyper-parameters; and 4) a demonstration of the real-world performance of Con-GAE. 



\subsection{Dataset}
\textit{Uber Movement} dataset~\cite{UM} provides multiple publicly available mobility datasets including origin-destination travel times at an hourly basis. We use data from the first two quarters of 2019 (Jan. - June) in Nashville, TN for analysis, which is the latest data available. The Q2 data is used for training, with the notable exclusion of data from April 15 - May 1, which is excluded because of a known large event, namely the NFL draft, occurred in Nashville on April 25-27. The Q1 data is used for testing, and the three-week period containing the NFL draft is used for a qualitative case study.


\textit{New York City (NYC) taxi dataset}~\cite{NYC} is an open source dataset recording pick-up and drop-off time and locations for each taxi trip in NYC. We aggregate the trip information into origin-destination travel time matrices for taxi zones at an hourly basis. The data from Jan.1 - Mar.31, 2019 is used for training, and Apr.1 - June.30 2019 is used for testing. 

\textit{Chicago taxi dataset}~\cite{taxi} records the taxi trips in the City of Chicago, and is processed in the same way as NYC taxi dataset to result in origin-destination travel time matrices at an hourly basis. The data from Jan.1 - Mar.31 2019 is used for training, and Apr.1 - Apr.30, 2019 is used for testing. 

Table~\ref{table:stats} shows the statistics of the three datasets. The plot for travel time distribution can be found in appendix. For all three datasets, we restrict our analysis to the top 50 zones, covering at least 75\% of all data. because no ground truth anomalies are available, synthetic events are added to testing sets, following a similar anomaly injection approach of~\cite{akoglu2015graph, yu2018netwalk}, 


\begin{table}
\centering
\caption{Dataset statistics}
\label{table:stats}

\resizebox{.95\columnwidth}{!}{
\begin{tabular}{l|cccccc} 
\toprule
Dataset  & Uber Movement & NYC taxi & Chicago taxi  \\ 
\hline
\# graph samples  & 4331 &  4344 & 2880 \\
avg travel time (min) & 5.3 &14.8&13.4\\
avg \# edges per graph & 1471  &1482 &440\\
\bottomrule
\end{tabular}}
\end{table}

\subsection{Baselines}
We compare our method against both traditional and state-of-the-art models for anomaly detection spanning multiple detection strategies. The baselines are: \textit{i}) A naive \textit{historical average} (HA) approach which calculates an anomaly score as the mean squared error between all observed OD pairs with the historical mean value at the corresponding day of week and hour of day, thereby capturing periodicity of the data: \textit{ii}) A \textit{Robust tensor recovery and completion} (RTC)~\cite{hu2019robust} method that uses a low-rank tensor decomposition to capture spatial-temporal correlations and detect outliers; \textit{iii}) An autoencoder~\cite{hawkins2002outlier} using MLP for the encoder and decoder; \textit{iv}) the 
\textit{Long Short Term Memory Networks based Encoder-Decoder scheme for Anomaly Detection} (EncDec-AD)~\cite{malhotra2016lstm}; \textit{v}) a \textit{Deep autoencoding Gaussian mixture model} (DAGMM)~\cite{zong2018deep}; \textit{vi}) A RNN-based~\textit{deep structured energy based model} REBM~\cite{zhai2016deep}; and two graph-based baselines: \textit{vii)} \textit{GCN}~\cite{kipf2017semi}; \textit{viii)} GraphSAGE~\cite{hamilton2017inductive}. The graph-based methods GCN and GraphSAGE have an encoder-decoder structure to encode each node in a low dimensional space, thus can serve as an autoencoder for anomaly detection. A description of each method can be found in supplementary material.

\subsection{Experiment setup}
We consider two kinds of anomalies: \textit{i}) spatial anomalies, where the traffic conditions  deviate from normal spatial patterns, and \textit{ii}) temporal anomalies,  where traffic conditions follow a correct spatial pattern, but is not the typical condition for the corresponding time of day.

Spatial anomalies are injected by first randomly selecting a fraction $\gamma$ of time slices to pollute. For each polluted time slice, we randomly choose a fraction $\alpha$ of the OD pairs, and perturb the corresponding travel time by a factor drawn from an uniform distribution $\mathcal{U}(-\beta, \beta)$. 
Temporal anomalies are introduced by randomly selecting a fraction $\gamma$ of time slices, and shifting the time associated with the data by 12 hours (e.g., 8pm becomes 8am and vise versa).

In the experiments, the pollution ratio and magnitudes are set to reflect the assumption that large events are infrequent, but they pollute relatively large portions of the network. We also vary  the pollution ratios and magnitudes $\alpha, \beta$, and $\gamma$ to check the anomaly detection performance under various cases. For each case, we randomly generate the test set five times and calculate the average model performance.

The \textit{area under the receiver operating characteristic curve} (AUC) is used as the metric to compare the methods. For the deep learning methods (i.e., Con-GAE, AE, DAGMM, EncDec-AD, REBM, GCN and GraphSAGE), the output of a given sample is an anomaly score. Using the anomaly scores of the test samples, the AUC is computed. For RTC, the output includes a sparse tensor containing anomalies. A anomaly score is calculated using the $l_2$ norm of the vector corresponding to each sample.

\subsection{Model Configurations}

The model settings for Con-GAE is as follows. The node feature matrix $X \in \mathbb{R}^{N \times d}$ is constructed with $d = 4$ corresponding to the minimum and maximum latitude and longitude extents of the zone corresponding to the node.
The encoder uses $L=2$ layers of GCN to learn the node-level embedding. The model is trained for 150 epochs with a batch size of 10. Out of the training set, 10\% is kept out as validation set for early stopping. Other settings of Con-GAE for different datasets are as follows.

For Uber data, the two layers of node embedding $h_i^1(t)\in \mathbb{R}^{300}$ and  $h_i^2(t)\in \mathbb{R}^{150}$ respectively. The dimension of the hour and week embeddings are $d_\text{week}=d_\text{hour} = 100$, and the graph embedding dimension is $d_g = 150$. The dropout rates $p_\text{e\_drop}$ and $p_\text{drop}$ are both set at 0.2. The initial learning rate of the Adam optimizer is $5\times 10^{-5}$, and we decay the learning rate by 0.5 every 50 epochs. 

For NYC data, the two layers of node embedding $h_i^1(t)\in \mathbb{R}^{150}$ and  $h_i^2(t)\in \mathbb{R}^{50}$ respectively. The dimension of the hour and week embeddings are $d_\text{week}=d_\text{hour} = 100$, and the graph embedding dimension is $d_g = 50$. The dropout rates $p_\text{e\_drop}$ and $p_\text{drop}$ are both set at 0.2. The initial learning rate is $1\times 10^{-3}$, and we decay the learning rate by 0.5 every 20 epochs. 

For Chicago data, the two layers of node embedding $h_i^1(t)\in \mathbb{R}^{300}$ and  $h_i^2(t)\in \mathbb{R}^{25}$ respectively. The dimension of the hour and week embeddings are $d_\text{week}=d_\text{hour} = 200$, and the graph embedding dimension is $d_g = 25$. The dropout rates $p_\text{e\_drop}$ and $p_\text{drop}$ are both set at 0.1. The initial learning rate is $1\times 10^{-3}$, and we decay the learning rate by 0.5 every 20 epochs. 


The AE, DAGMM, EncDec-AD, and REBM are implemented based on code in~\cite{time_series}. the respective hyper-parameters for each model are tuned to archive the best performance. GCN and GraphSAGE have the same encoding node dimension as our model. Detailed configurations are available in the supplementary materials.

\subsection{Result and analysis}

\begin{table}
\centering
\caption{ The AUC score for spatial and temporal anomaly detection of different datasets. }
\label{table:compare_datasets}
            
    \sisetup{detect-weight,mode=text}
    \renewrobustcmd{\bfseries}{\fontseries{b}\selectfont}
    \renewrobustcmd{\boldmath}{}
    \newrobustcmd{\B}{\bfseries}
    \addtolength{\tabcolsep}{-4.1pt}
\resizebox{.95\columnwidth}{!}{
\begin{tabular}{l|ccc|ccc} 
\toprule
  \multicolumn{1}{l|}{anomaly type}                      & \multicolumn{3}{c|}{spatial  anomaly} & \multicolumn{3}{c}{temporal anomaly}  \\ 
\cmidrule{1-1} \cmidrule{2-4}\cmidrule{5-7}
dataset  & Uber  & NYC & Chicago    & Uber     & NYC & Chicago  \\ 
\hline
HA        & 0.687  & 0.498  & 0.570   & 0.661  & 0.677  & 0.658   \\
RTC       & 0.765  & 0.795  & 0.601   & 0.522  & 0.247  & 0.562   \\
AE        & 0.812  & 0.671  & 0.727   & 0.517  & 0.592  & 0.460   \\
EncDec-AD & 0.582  & 0.782  & 0.715   & 0.549  & 0.659  & 0.671   \\
REBM      & 0.859  & 0.759  & 0.739   & 0.482  & 0.805  & 0.482   \\
DAGMM     & 0.546  & 0.603  & 0.669   & 0.466  & 0.397  & 0.414   \\
GraphSAGE & 0.840  & 0.741  & 0.674   & 0.542  & 0.464  & 0.426   \\
GCN       & 0.717  & 0.593  & 0.765   & 0.548  & 0.416  & 0.421   \\
\B Con-GAE   & \B 0.908  & \B 0.837  & \B 0.912   &  \B0.726  & \B 0.895  & \B 0.753   \\
\bottomrule
\end{tabular}}
\end{table}

\begin{table}
\centering
\caption{ The AUC score for spatial and temporal anomaly detection, under different anomaly rates. We vary the fraction $\gamma$ of the time slices chosen to be polluted.  }
\label{table:compare_st}
            
    \sisetup{detect-weight,mode=text}
    \renewrobustcmd{\bfseries}{\fontseries{b}\selectfont}
    \renewrobustcmd{\boldmath}{}
    \newrobustcmd{\B}{\bfseries}
    \addtolength{\tabcolsep}{-4.1pt}
\resizebox{.95\columnwidth}{!}{
\begin{tabular}{l|ccc|ccc} 
\toprule
  \multicolumn{1}{l|}{anomaly type}                      & \multicolumn{3}{c|}{spatial  anomaly} & \multicolumn{3}{c}{temporal anomaly}  \\ 
\cmidrule{1-1} \cmidrule{2-4}\cmidrule{5-7}
anomaly rate $\gamma$    & 5\%          & 10\% & 20\%     & 5\%          & 10\% & 20\%     \\ 
\hline
HA      &      0.728 &      0.687 &      0.711 &      0.658 &      0.661 &      0.659 \\
RTC     &      0.736 &      0.765 &      0.790 &      0.579 &      0.522 &      0.509 \\
AE      &      0.813 &      0.812 &      0.806 &      0.497 &      0.517 &      0.518 \\
EncDec-AD  &      0.584 &      0.582 &      0.582 &      0.564 &      0.549 &      0.532 \\
REBM    &      0.844 &      0.859 &      0.833 &      0.468 &      0.482 &      0.501 \\
DAGMM   &      0.550 &      0.546 &      0.507 &      0.439 &      0.466 &      0.477 \\
GraphSAGE  &      0.842 &  0.840 & 0.860 & 0.570 & 0.542 & 0.526 \\
GCN & 0.708 & 0.717 & 0.744 & 0.560 & 0.548 & 0.526\\

\B Con-GAE     &   \B   0.903 &    \B   0.908 &    \B   0.913 &   \B    0.752 &    \B   0.726 &    \B   0.693 \\
\bottomrule
\end{tabular}}
\end{table}

\subsubsection{Model comparison}
First, we compare Con-GAE  with baseline methods for both both spatial and temporal anomaly detection, for different datasets. We fix the fraction of the time slices chosen to be polluted $\gamma = 10\%$, pollution magnitude $\alpha = 50\%$ and $\beta = 10\%$. 
The results shown in Table~\ref{table:compare_datasets}. We can see that \GNN~constantly outperforms the other methods, with an improvement in AUC score between 0.1 and 0.4. For spatial anomalies, \GNN~is the only method to achieve an AUC above 0.9 for Uber data and Chicago data, and the only method above 0.8 for NYC data. For temporal anomalies, most other methods have AUC score around 0.5 except HA, meaning they are not well suited for detecting temporal anomalies, while \GNN~achieves an AUC score of 0.7 or higher.  These experiments highlight that when dealing with large scale mobility data, taking the graph structure into account can substantially boost the performance. Meanwhile, directly applying GCN and GraphSAGE model results in poor performance, showing the importance of designing hierarchical structure for graph-scale embedding beyond individual node embeddings when detecting network-wide anomalies. The temporal experiments shows that for data with periodicity, it is more effective to consider the long-term periodicity as in \GNN~and HA, than to consider the short-term dependencies as in LSTM-ED and REBM, or no temporal dependencies as in DAGMM, AE, GCN and GraphSAGE.

Then, we compare the anomaly detection methods under different  anomaly rates.  We vary the fraction $\gamma$ of the time slices chosen to be polluted, fixing $\alpha = 50\%$ and $\beta = 10\%$. The result is shown in Table~\ref{table:compare_st}. We can see that the advantage of our method holds under different anomaly rates.

Next, we investigate the influence of anomaly magnitude. We focus on spatial anomaly detection, and vary the fraction $\alpha$ of OD pairs polluted and the uniform distribution $\mathcal{U}(-\beta, \beta)$ deciding the range of pollution ratio. We fix the time slices polluted at $\gamma = 10\%$.  The result is shown in Table~\ref{table:compare_sp}. No method obtains the best performance over all scenarios. All methods have higher AUC score with larger $\alpha$ and $\beta$, and perform less well when the pollution is more nuanced. \GNN~has significant advantage over the other methods for $\beta=10\%$ or larger. When $\beta$ is lower than $10\%$, only the linear method RTC performs adequately well. Since our goal is to detect large events with significant disturbance to road networks, we conclude that \GNN~is the most competitive method  overall.

\begin{table}
\centering
\caption{The AUC score for spatial anomaly detection, under different magnitudes of anomalies. Results are shown under different $\alpha$ deciding the OD pairs polluted, and different $\beta$ corresponding to the perturbation magnitude.}
\label{table:compare_sp}
            
    \sisetup{detect-weight,mode=text}
    \renewrobustcmd{\bfseries}{\fontseries{b}\selectfont}
    \renewrobustcmd{\boldmath}{}
    \newrobustcmd{\B}{\bfseries}
    \addtolength{\tabcolsep}{-4.1pt}
\resizebox{0.95\columnwidth}{!}{
\begin{tabular}{p{0.3\columnwidth}|ccc|ccc} 
\toprule
  \multicolumn{1}{p{0.12\columnwidth}|}{spatial anomaly rate $\alpha$}    & \multicolumn{3}{c|}{25\%} & \multicolumn{3}{c}{50\%}  \\ 
\cmidrule{1-1} \cmidrule{2-4}\cmidrule{5-7}
 anomaly magnitude $\beta$    & 5\%          & 10\% & 20\%     & 5\%          & 10\% & 20\%     \\ 
\hline
HA      &      0.405 &      0.533 &      0.804 &      0.455 &      0.687 &      0.934 \\
RTC     &     \B 0.626 &      0.699 &      0.863 &   \B   0.648 &      0.765 &      0.942 \\
AE      &      0.294 &      0.572 &      0.936 &      0.405 &      0.812 &      0.994 \\
EncDec-AD  &      0.410 &      0.483 &      0.727 &      0.452 &      0.582 &      0.896 \\
REBM    &      0.389 &      0.633 &      0.958 &      0.491 &      0.859 &      0.997 \\
DAGMM   &      0.511 &      0.527 &      0.567 &      0.525 &      0.546 &      0.639 \\
GraphSAGE  &  0.381 &  0.627 &  0.963 &  0.491 &  0.840 &  1.000        \\
GCN & 0.443 & 0.564 & 0.844 & 0.498 & 0.717 & 0.966 \\
\B Con-GAE     &      0.482 &    \B  0.755 &  \B    0.985 &      0.610 &    \B  0.908 &   \B   1.000 \\
\bottomrule
\end{tabular}}
\end{table}

\subsubsection{Ablation study}
We compare the several variants of Con-GAE, to see how each component of \GNN~helps with anomaly detection. We consider the following variants: 
\textit{i}) \textit{Con-GAE-sp}: We remove all temporal information the autoencoder, and only use the spatial graph data for training and detection.  
\textit{ii}) \textit{Con-GAE-t}: we remove the GCN layers, and only use the temporal information for training and detection.
\textit{iii}) \textit{Con-GAE-fc}: we replace the GCN layers in the autoencoder with fully connected layers, while temporal information is retained. \textit{iv}) \textit{Con-GAE-NonContextDec}: we omit the context embedding and only use the graph embedding as input to the decoder. \textit{v)} \textit{Con-GAE-NonWeightedEnc}: We use the standard GraphSAGE implementation~\eqref{eq:SAGE} with mean aggregator as first step of encoder, instead of the modified weighted mean aggregator.

The result is shown in Table~\ref{table:ablation}. We can see that the modified GCN layers play a crucial role in anomaly detection. The AUC score is 0.2 lower if we remove the GCN layers (Con-GAE-t); and is 0.1 lower if we replace GCN layers with fully connected layers (Con-GAE-fc), or with standard GraphSAGE layers (Con-GAE-NonWeightedEnc). We observe that temporal information provides marginal benefit for spatial anomaly detection, but is crucial for temporal anomaly detection. Without temporal information, it is impossible for Con-GAE-sp to detect temporal anomalies, with an AUC score constantly around 0.5. The drop of performance in Con-GAE-NonContextDec shows that explicitly conditioning the decoder on time context helps the decoder to better reconstruct the graph, even though the temporal information might be encoded by the encoder. \Yue{delete the Con-GAE-NonContextDec? }

\begin{table}
\centering
\caption{Ablation study.}
\label{table:ablation}
            
    \sisetup{detect-weight,mode=text}
    \renewrobustcmd{\bfseries}{\fontseries{b}\selectfont}
    \renewrobustcmd{\boldmath}{}
    \newrobustcmd{\B}{\bfseries}
    \addtolength{\tabcolsep}{-4.1pt}
\resizebox{.95\columnwidth}{!}{
\begin{tabular}{l|ccc|ccc} 
\toprule
  \multicolumn{1}{l|}{anomaly type}                      & \multicolumn{3}{c|}{spatial  anomaly} & \multicolumn{3}{c}{temporal anomaly}  \\ 
\cmidrule{1-1} \cmidrule{2-4}\cmidrule{5-7}
anomaly rate $\gamma$    & 5\%          & 10\% & 20\%     & 5\%          & 10\% & 20\%     \\ 
\hline
Con-GAE     &    \B  0.903 &  \B    0.908 &    \B  0.913 &  \B    0.752 &    \B  0.726 &   \B   0.693 \\
Con-GAE-t  &      0.754 &      0.722 &      0.729 &      0.630 &      0.630 &      0.621 \\
Con-GAE-sp  &      0.888 &      0.893 &      0.905 &      0.572 &      0.549 &      0.529 \\
Con-GAE-fc &      0.846 &      0.844 &      0.829 &      0.693 &      0.681 &      0.683 \\
Con-GAE-NonContextDec & 0.829 & 0.816 & 0.818 & 0.544&  0.541 &  0.538 \\
Con-GAE-NonWeightedEnc & 0.835 & 0.813 & 0.819 & 0.606&  0.591 &0.576\\
\bottomrule
\end{tabular}}
\end{table}

\subsubsection{Sensitivity analysis}
Next, we explore how the performance of~\GNN~changes as a function of the node embedding dimension, the hour and week embedding dimension, the low dimensional final graph embedding, and the model depth. We fix $\alpha = 50\%, \beta = 10\% $ and $ \gamma = 10\%$. 

\begin{figure}
\centering

\begin{subfigure}[t]{0.48\columnwidth}
    \includegraphics[width= \linewidth]{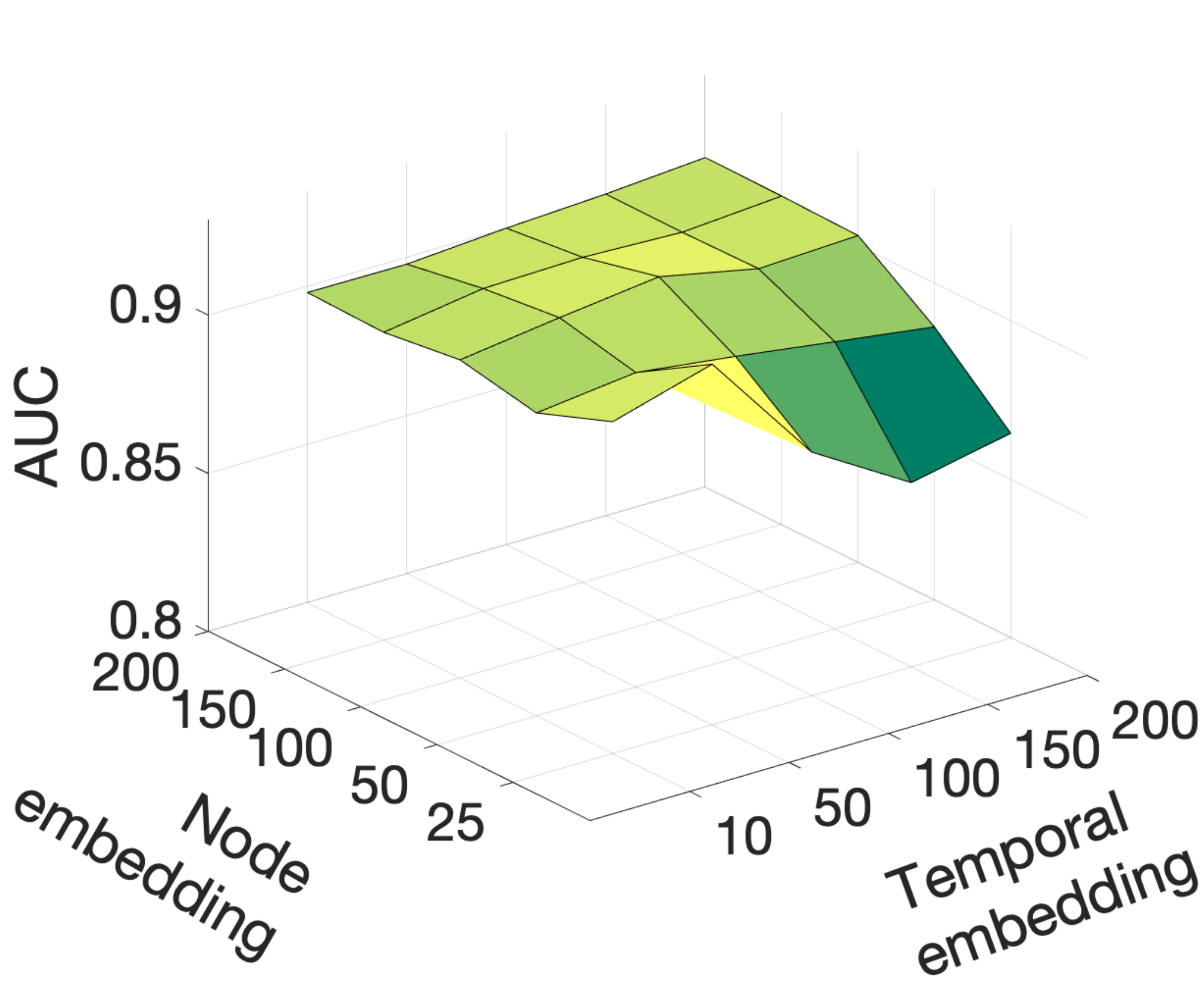}
    \caption{AUC score of Con-GAE with varying dimension of node and time embedding.}
    \label{fig:sens_3D}
\end{subfigure}
\quad
\begin{subfigure}[t]{0.46\columnwidth}
    \includegraphics[width= \linewidth]{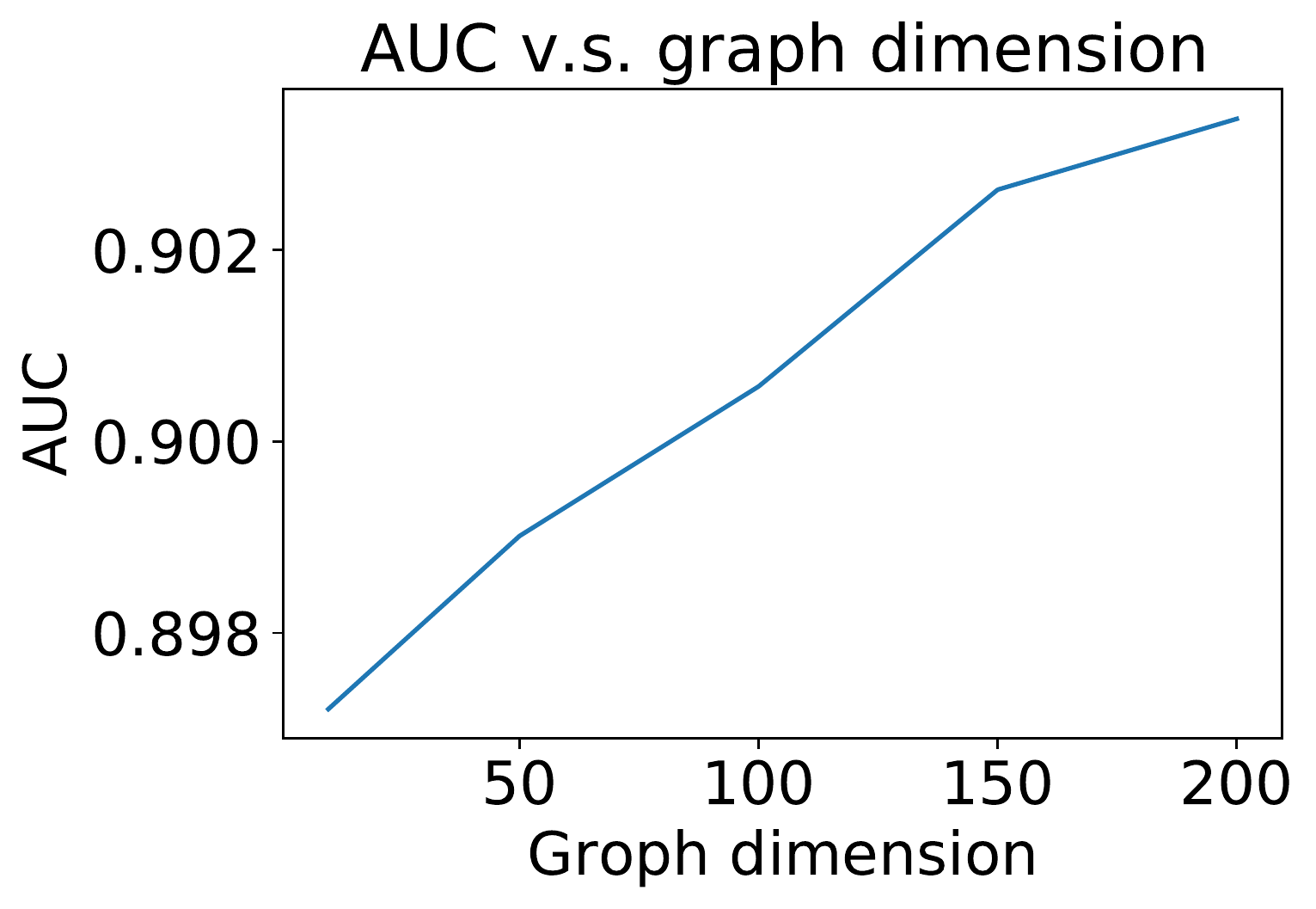}
    \caption{AUC score of Con-GAE with varying graph embedding dimension.}
    \label{fig:sens_hidden}
\end{subfigure}
\caption{Sensitivity analysis of Con-GAE }
\label{fig:AUCscore}
\end{figure}

First, we explore the sensitivity to spatial and temporal encoding dimensions. We vary the node embedding dimension between 25 and 200, and vary the week and hour embedding dimension between 10 and 200.
We fix the other parameters the same as in Model Configurations section. The result is shown in Fig~\ref{fig:sens_3D}. We can see that~\GNN~is not too sensitive to spatial encoding dimension and temporal embedding dimension, with the AUC all in the range of 0.84-0.94. Most combinations work well except the case when spatial encoding dimension is too low (below 50) and the temporal embedding dimension is too high (above 100). 

\begin{table}
\centering
\caption{Sensitivity to model depth}
\label{table:depth_sens}

\resizebox{.95\columnwidth}{!}{
\begin{tabular}{l|cccccc} 
\toprule
\# GCN layers  & 1    &2 & 3  & 4   &5 &6  \\ 
\hline
AUC score   &  0.896 &    0.908 & 0.901 &  0.896&0.886 &0.889 \\
\bottomrule
\end{tabular}}
\end{table}

We then fix node embedding and temporal embedding dimensions, and explore the influence of the final graph embedding dimension $d_g$. We vary graph dimension from 10 to 150. Fig.~\ref{fig:sens_hidden} shows the result. We can see that AUC score is insensitive with scores slightly increasing but all around 0.9.

Lastly, we investigate the sensitivity to model depth. We vary the number of GCN layers from 1 to 6, fixing  $h_i^1(t)\in \mathbb{R}^{300}$ and  $h_i^l(t)\in \mathbb{R}^{150}$ for layers $l >1$. The result is shown in Table~\ref{table:depth_sens}. We can see that \GNN~is not sensitive to depth, with AUC score all between 0.88 and 0.91.


\subsection{Visualization of real-world traffic anomaly}
In this section, we demonstrate the real-world application of \GNN. We use the test set 4/27-5/01 of 2019 Q2, which is not seen during training and is known to contain a large event (the NFL draft). The NFL draft brought several hundred thousand visitors to the city, and nightly events resulted in roadway closures in the downtown Nashville area. 

\begin{figure}
    \centering
    \includegraphics[width= 0.9\linewidth]{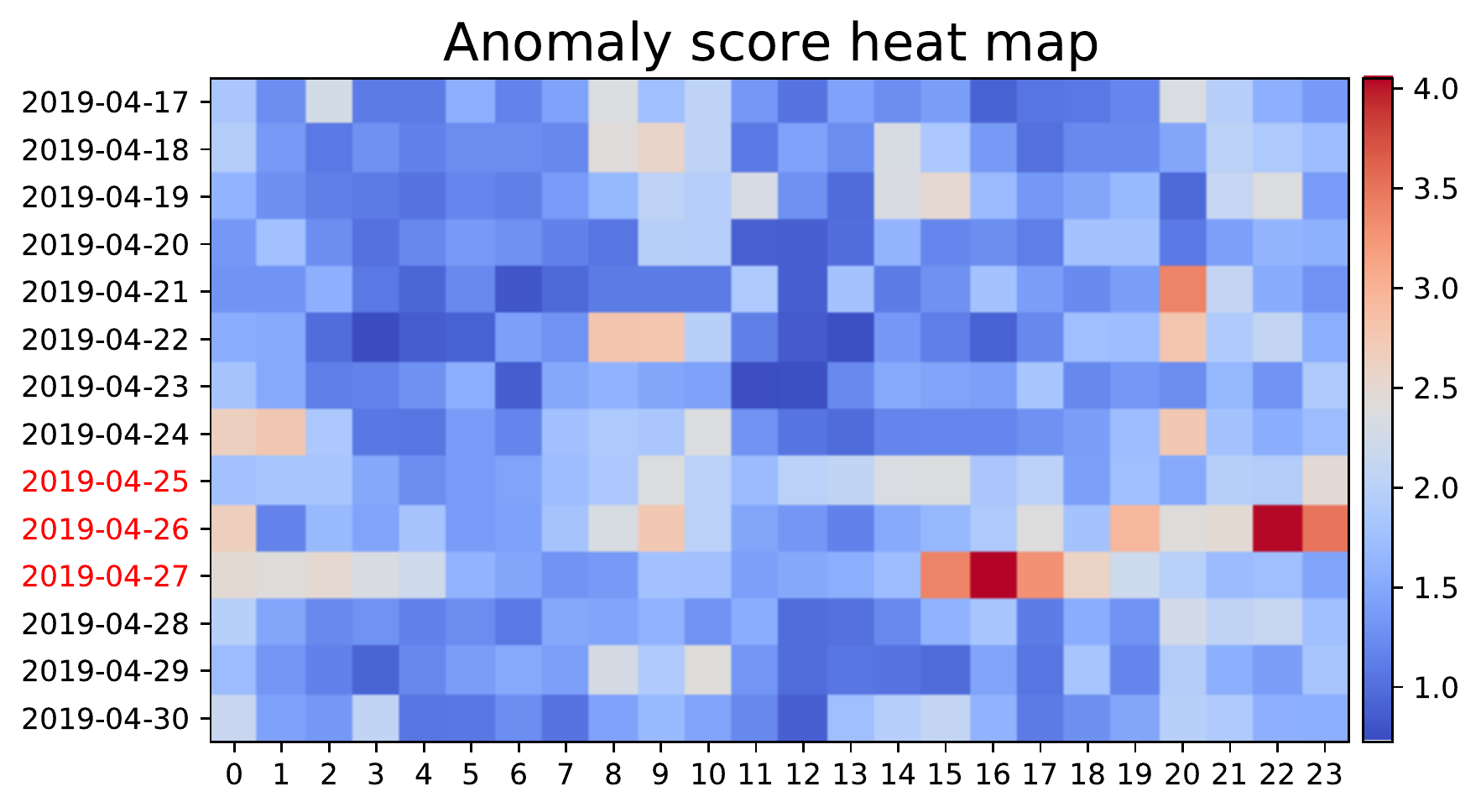}
    \caption{Anomaly sores for each hour. The NFL days are marked red. The score denotes the reconstruction error~\eqref{eq:MSE}.}
    \label{fig:NFL}
\end{figure}

Figure~\ref{fig:NFL} shows the anomaly score (the reconstruction loss) for each hour of the studied period. The days under which the major activities of the draft take place are highlighted in red on the y-axis. We can see that the highest anomaly scores occur during the draft, specifically 4/26 19:00 until midnight, as well as 4/27 15:00-18:00. These are the time when the draft, post draft and post concert \& fireworks take place~\cite{NFL2019}. 

\begin{figure*}
    \centering
    \includegraphics[width= 0.85\linewidth]{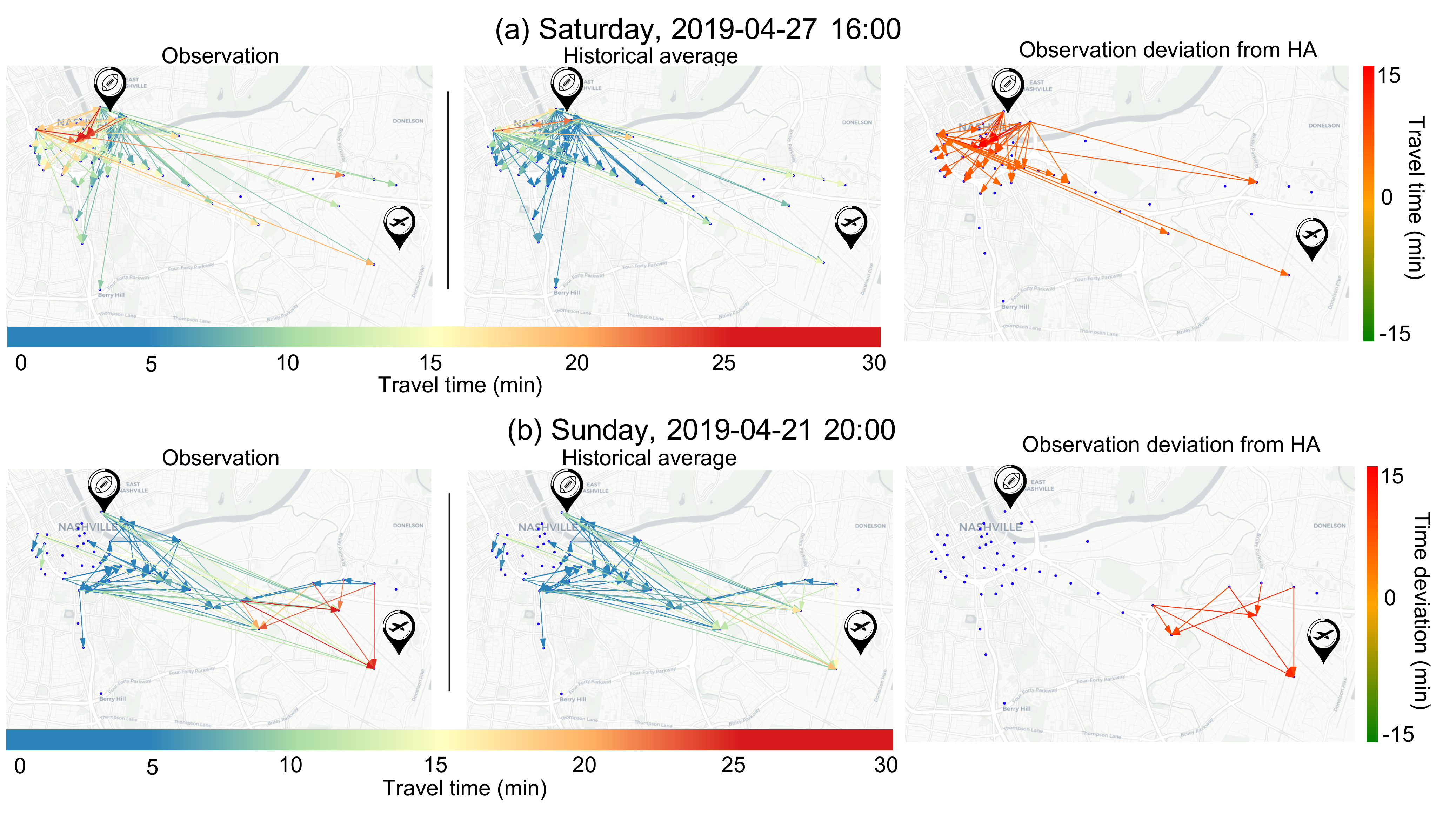}
    \caption{Traffic condition for the detected events. The OD pair travel time and comparison to historical conditions are plotted.}
    \label{fig:map_tt}
\end{figure*}

Figure~\ref{fig:map_tt} plots the OD travel time data during two periods with high anomaly scores. For clarity, only a subset of travel time pairs are plotted, and only deviations in excess of 5 min from the historical average are shown. Fig~\ref{fig:map_tt}(a) shows that that the travel times from the region containing the draft to other parts of the city is significantly longer than usual.
The third largest anomaly period occurs in the consecutive days of 4/21 to 4/24 at 20:00. This matches the record of \textit{Tennessee Department of Transportation}~\cite{TDOT2019} that there are alternating nightly maintenance road closures for the major freeway connecting downtown to the airport, resulting in substantially longer travel times (Fig~\ref{fig:map_tt}(b)).

\section{Related Work}
We review graph network embedding and autoencoder-based anomaly detection techniques that inspire Con-GAE.
\subsection{Graph network embedding}
In recent years, there has been rapid development in graph embedding-based learning algorithms.
In particular, GCN has gained popularity because of its scalability and good performance~\cite{dwivedi2020benchmarking}. Some benchmark variations of GCN  include~\cite{kipf2017semi, hamilton2017inductive,velivckovic2017graph}, which have become the building block of many complex graph network models~\cite{wu2020comprehensive}. Our work focuses on weighted directed graphs appropriate for mobility datasets, compared with binary undirected graphs in the original work of~\cite{kipf2017semi, hamilton2017inductive}.

A \textit{graph autoencoder} (GAE) is an unsupervised learning approach that encodes the graph in a latent space, and reconstructs the graph structure from the encoding~\cite{wu2020comprehensive}. 
Since the introduction of GAE in~\cite{kipf2016variational, tian2014learning}, it has been widely used, for example in link prediction~\cite{wang2019robust, schlichtkrull2018modeling,yang2014embedding}, matrix completion~\cite{berg2017graph}, and graph clustering~\cite{wang2017mgae}.
While existing GAE~\cite{kipf2016variational} learn node embeddings during encoding, our model further compresses the node embeddings of all nodes in a graph to a single graph embedding, and conditions it on temporal information to add additional context important for interpreting mobility data.

Given the intrinsic graph structure of transportation networks, recent research has demonstrated advantages in applying graph embedding on mobility data, for forecasting\cite{yu2017spatio, diao2019dynamic}, passenger demand prediction~\cite{wang2019origin}, multi-modal transportation recommendation~\cite{liu2019joint}, and human trajectory prediction~\cite{mohamed2020social}. While graph embedding for traffic anomaly detection is less studied, our present work work we show the ability of graph embedding to learn patterns in OD mobility data for anomaly detection.

\subsection{Autoencoder-based anomaly detection}
Deep autoencoder based methods have been widely used for anomaly detection~\cite{zhou2017anomaly,chong2017abnormal, xu2018unsupervised, borghesi2019anomaly, malhotra2016lstm, sakurada2014anomaly}.
The encoder and decoder can have various structures based on different input data types. Besides the widely used MLP based encoder and decoders~\cite{xu2018unsupervised, borghesi2019anomaly, sakurada2014anomaly},  \textit{convolutional neural networks} (CNN) are applied on grid data~\cite{chong2017abnormal}, and \textit{recurrent neural networks} (RNN) are applied on time series data~\cite{malhotra2016lstm,su2019robust}.
Recently, GCN-based autoencoders have been used for anomaly detection on graph data~\cite{li2019specae,ding2019deep, kumagai2020semi, kulkarni2020deep,zhu2020deepad,fan2020anomalydae}, with an important emphasis node-level anomalies. In comparison, our work aims at graph-level anomaly detection, since each graph corresponds to the network wide traffic condition at a specific time instant that may be disturbed by a large scale event.


\section{Conclusion}
In this work, we address the problem of detecting large events in large origin destination mobility datasets. We formulate the problem of detecting these events as a problem of detecting anomalies in a set of time dependent directed graphs containing mobility data on the network. We introduced~\GNN which is an autoencoder based anomaly detector that uses GCN layers and temporal embeddings to determine anomalies.  Extensive experiments on synthetic test sets generated from real-world mobility data shows that \GNN~outperforms several state-of-art anomaly detection methods in both spatial anomaly and temporal detection. In future work, we are interested to explore scalability of~\GNN~to networks with significantly more nodes, and for anomaly detection on other network datasets.


\bibliographystyle{ACM-Reference-Format}
\bibliography{sample-base}


\begin{thebibliography}{51}


\ifx \showCODEN    \undefined \def \showCODEN     #1{\unskip}     \fi
\ifx \showDOI      \undefined \def \showDOI       #1{#1}\fi
\ifx \showISBNx    \undefined \def \showISBNx     #1{\unskip}     \fi
\ifx \showISBNxiii \undefined \def \showISBNxiii  #1{\unskip}     \fi
\ifx \showISSN     \undefined \def \showISSN      #1{\unskip}     \fi
\ifx \showLCCN     \undefined \def \showLCCN      #1{\unskip}     \fi
\ifx \shownote     \undefined \def \shownote      #1{#1}          \fi
\ifx \showarticletitle \undefined \def \showarticletitle #1{#1}   \fi
\ifx \showURL      \undefined \def \showURL       {\relax}        \fi
\providecommand\bibfield[2]{#2}
\providecommand\bibinfo[2]{#2}
\providecommand\natexlab[1]{#1}
\providecommand\showeprint[2][]{arXiv:#2}

\bibitem[\protect\citeauthoryear{Akoglu, Tong, and Koutra}{Akoglu
  et~al\mbox{.}}{2015}]%
        {akoglu2015graph}
\bibfield{author}{\bibinfo{person}{Leman Akoglu}, \bibinfo{person}{Hanghang
  Tong}, {and} \bibinfo{person}{Danai Koutra}.}
  \bibinfo{year}{2015}\natexlab{}.
\newblock \showarticletitle{Graph based anomaly detection and description: a
  survey}.
\newblock \bibinfo{journal}{\emph{Data mining and knowledge discovery}}
  \bibinfo{volume}{29}, \bibinfo{number}{3} (\bibinfo{year}{2015}),
  \bibinfo{pages}{626--688}.
\newblock


\bibitem[\protect\citeauthoryear{Berg, Kipf, and Welling}{Berg
  et~al\mbox{.}}{2018}]%
        {berg2017graph}
\bibfield{author}{\bibinfo{person}{Rianne van~den Berg},
  \bibinfo{person}{Thomas~N Kipf}, {and} \bibinfo{person}{Max Welling}.}
  \bibinfo{year}{2018}\natexlab{}.
\newblock \showarticletitle{Graph convolutional matrix completion}.
\newblock \bibinfo{journal}{\emph{KDD’18 Deep Learning Day}}
  (\bibinfo{year}{2018}).
\newblock


\bibitem[\protect\citeauthoryear{Borghesi, Bartolini, Lombardi, Milano, and
  Benini}{Borghesi et~al\mbox{.}}{2019}]%
        {borghesi2019anomaly}
\bibfield{author}{\bibinfo{person}{Andrea Borghesi}, \bibinfo{person}{Andrea
  Bartolini}, \bibinfo{person}{Michele Lombardi}, \bibinfo{person}{Michela
  Milano}, {and} \bibinfo{person}{Luca Benini}.}
  \bibinfo{year}{2019}\natexlab{}.
\newblock \showarticletitle{Anomaly detection using autoencoders in high
  performance computing systems}. In \bibinfo{booktitle}{\emph{Proceedings of
  the AAAI Conference on Artificial Intelligence}}, Vol.~\bibinfo{volume}{33}.
  \bibinfo{pages}{9428--9433}.
\newblock


\bibitem[\protect\citeauthoryear{{Chicago}}{{Chicago}}{2020a}]%
        {scooter}
\bibfield{author}{\bibinfo{person}{{Chicago}}.}
  \bibinfo{year}{2020}\natexlab{a}.
\newblock \bibinfo{title}{Divvy Trips}.
\newblock
  \bibinfo{howpublished}{\url{https://data.cityofchicago.org/Transportation/Divvy-Trips/fg6s-gzvg}}.
\newblock


\bibitem[\protect\citeauthoryear{{Chicago}}{{Chicago}}{2020b}]%
        {taxi}
\bibfield{author}{\bibinfo{person}{{Chicago}}.}
  \bibinfo{year}{2020}\natexlab{b}.
\newblock \bibinfo{title}{Taxi Trips}.
\newblock
  \bibinfo{howpublished}{\url{https://data.cityofchicago.org/Transportation/Taxi-Trips/wrvz-psew}}.
\newblock


\bibitem[\protect\citeauthoryear{Chong and Tay}{Chong and Tay}{2017}]%
        {chong2017abnormal}
\bibfield{author}{\bibinfo{person}{Yong~Shean Chong} {and}
  \bibinfo{person}{Yong~Haur Tay}.} \bibinfo{year}{2017}\natexlab{}.
\newblock \showarticletitle{Abnormal event detection in videos using
  spatiotemporal autoencoder}. In \bibinfo{booktitle}{\emph{International
  Symposium on Neural Networks}}. Springer, \bibinfo{pages}{189--196}.
\newblock


\bibitem[\protect\citeauthoryear{Diao, Wang, Zhang, Liu, Xie, and He}{Diao
  et~al\mbox{.}}{2019}]%
        {diao2019dynamic}
\bibfield{author}{\bibinfo{person}{Zulong Diao}, \bibinfo{person}{Xin Wang},
  \bibinfo{person}{Dafang Zhang}, \bibinfo{person}{Yingru Liu},
  \bibinfo{person}{Kun Xie}, {and} \bibinfo{person}{Shaoyao He}.}
  \bibinfo{year}{2019}\natexlab{}.
\newblock \showarticletitle{Dynamic spatial-temporal graph convolutional neural
  networks for traffic forecasting}. In \bibinfo{booktitle}{\emph{Proceedings
  of the AAAI Conference on Artificial Intelligence}},
  Vol.~\bibinfo{volume}{33}. \bibinfo{pages}{890--897}.
\newblock


\bibitem[\protect\citeauthoryear{Ding, Li, Bhanushali, and Liu}{Ding
  et~al\mbox{.}}{2019}]%
        {ding2019deep}
\bibfield{author}{\bibinfo{person}{Kaize Ding}, \bibinfo{person}{Jundong Li},
  \bibinfo{person}{Rohit Bhanushali}, {and} \bibinfo{person}{Huan Liu}.}
  \bibinfo{year}{2019}\natexlab{}.
\newblock \showarticletitle{Deep anomaly detection on attributed networks}. In
  \bibinfo{booktitle}{\emph{Proceedings of the 2019 SIAM International
  Conference on Data Mining}}. SIAM, \bibinfo{pages}{594--602}.
\newblock


\bibitem[\protect\citeauthoryear{Dwivedi, Joshi, Laurent, Bengio, and
  Bresson}{Dwivedi et~al\mbox{.}}{2020}]%
        {dwivedi2020benchmarking}
\bibfield{author}{\bibinfo{person}{Vijay~Prakash Dwivedi},
  \bibinfo{person}{Chaitanya~K Joshi}, \bibinfo{person}{Thomas Laurent},
  \bibinfo{person}{Yoshua Bengio}, {and} \bibinfo{person}{Xavier Bresson}.}
  \bibinfo{year}{2020}\natexlab{}.
\newblock \showarticletitle{Benchmarking graph neural networks}.
\newblock \bibinfo{journal}{\emph{arXiv preprint arXiv:2003.00982}}
  (\bibinfo{year}{2020}).
\newblock


\bibitem[\protect\citeauthoryear{Fan, Zhang, and Li}{Fan et~al\mbox{.}}{2020}]%
        {fan2020anomalydae}
\bibfield{author}{\bibinfo{person}{Haoyi Fan}, \bibinfo{person}{Fengbin Zhang},
  {and} \bibinfo{person}{Zuoyong Li}.} \bibinfo{year}{2020}\natexlab{}.
\newblock \showarticletitle{AnomalyDAE: Dual autoencoder for anomaly detection
  on attributed networks}. In \bibinfo{booktitle}{\emph{ICASSP 2020-2020 IEEE
  International Conference on Acoustics, Speech and Signal Processing
  (ICASSP)}}. IEEE, \bibinfo{pages}{5685--5689}.
\newblock


\bibitem[\protect\citeauthoryear{Fey and Lenssen}{Fey and Lenssen}{2019}]%
        {Fey/Lenssen/2019}
\bibfield{author}{\bibinfo{person}{Matthias Fey} {and} \bibinfo{person}{Jan~E.
  Lenssen}.} \bibinfo{year}{2019}\natexlab{}.
\newblock \showarticletitle{Fast Graph Representation Learning with {PyTorch
  Geometric}}. In \bibinfo{booktitle}{\emph{ICLR Workshop on Representation
  Learning on Graphs and Manifolds}}.
\newblock


\bibitem[\protect\citeauthoryear{Fischer, Gierke, Kellermeier, Kesar, Stebner,
  and Thevessen}{Fischer et~al\mbox{.}}{2019}]%
        {time_series}
\bibfield{author}{\bibinfo{person}{Maxi Fischer}, \bibinfo{person}{Willi
  Gierke}, \bibinfo{person}{Thomas Kellermeier}, \bibinfo{person}{Ajay Kesar},
  \bibinfo{person}{Axel Stebner}, {and} \bibinfo{person}{Daniel Thevessen}.}
  \bibinfo{year}{2019}\natexlab{}.
\newblock \bibinfo{title}{{Anomaly Detection on Time Series: An Evaluation of
  Deep Learning Methods}}.
\newblock
  \bibinfo{howpublished}{\url{https://github.com/KDD-OpenSource/DeepADoTS}}.
\newblock
\newblock
\shownote{Accessed: 2020 Jul.}


\bibitem[\protect\citeauthoryear{Goyal, Kamra, He, and Liu}{Goyal
  et~al\mbox{.}}{2018}]%
        {goyal2018dyngem}
\bibfield{author}{\bibinfo{person}{Palash Goyal}, \bibinfo{person}{Nitin
  Kamra}, \bibinfo{person}{Xinran He}, {and} \bibinfo{person}{Yan Liu}.}
  \bibinfo{year}{2018}\natexlab{}.
\newblock \showarticletitle{Dyngem: Deep embedding method for dynamic graphs}.
\newblock \bibinfo{journal}{\emph{arXiv preprint arXiv:1805.11273}}
  (\bibinfo{year}{2018}).
\newblock


\bibitem[\protect\citeauthoryear{Hamilton, Ying, and Leskovec}{Hamilton
  et~al\mbox{.}}{2017}]%
        {hamilton2017inductive}
\bibfield{author}{\bibinfo{person}{Will Hamilton}, \bibinfo{person}{Zhitao
  Ying}, {and} \bibinfo{person}{Jure Leskovec}.}
  \bibinfo{year}{2017}\natexlab{}.
\newblock \showarticletitle{Inductive representation learning on large graphs}.
  In \bibinfo{booktitle}{\emph{Advances in neural information processing
  systems}}. \bibinfo{pages}{1024--1034}.
\newblock


\bibitem[\protect\citeauthoryear{Hawkins, He, Williams, and Baxter}{Hawkins
  et~al\mbox{.}}{2002}]%
        {hawkins2002outlier}
\bibfield{author}{\bibinfo{person}{Simon Hawkins}, \bibinfo{person}{Hongxing
  He}, \bibinfo{person}{Graham Williams}, {and} \bibinfo{person}{Rohan
  Baxter}.} \bibinfo{year}{2002}\natexlab{}.
\newblock \showarticletitle{Outlier detection using replicator neural
  networks}. In \bibinfo{booktitle}{\emph{International Conference on Data
  Warehousing and Knowledge Discovery}}. Springer, \bibinfo{pages}{170--180}.
\newblock


\bibitem[\protect\citeauthoryear{Hu and Work}{Hu and Work}{2019}]%
        {hu2019robust}
\bibfield{author}{\bibinfo{person}{Yue Hu} {and} \bibinfo{person}{Dan Work}.}
  \bibinfo{year}{2019}\natexlab{}.
\newblock \showarticletitle{Robust Tensor Recovery with Fiber Outliers for
  Traffic Events}.
\newblock \bibinfo{journal}{\emph{arXiv preprint arXiv:1908.10198}}
  (\bibinfo{year}{2019}).
\newblock


\bibitem[\protect\citeauthoryear{Kingma and Ba}{Kingma and Ba}{2015}]%
        {kingma2014adam}
\bibfield{author}{\bibinfo{person}{Diederik~P Kingma} {and}
  \bibinfo{person}{Jimmy Ba}.} \bibinfo{year}{2015}\natexlab{}.
\newblock \showarticletitle{Adam: A method for stochastic optimization}.
\newblock \bibinfo{journal}{\emph{ICLR 2015}} (\bibinfo{year}{2015}).
\newblock


\bibitem[\protect\citeauthoryear{Kipf and Welling}{Kipf and Welling}{2016}]%
        {kipf2016variational}
\bibfield{author}{\bibinfo{person}{Thomas~N Kipf} {and} \bibinfo{person}{Max
  Welling}.} \bibinfo{year}{2016}\natexlab{}.
\newblock \showarticletitle{Variational graph auto-encoders}.
\newblock \bibinfo{journal}{\emph{arXiv preprint arXiv:1611.07308}}
  (\bibinfo{year}{2016}).
\newblock


\bibitem[\protect\citeauthoryear{Kipf and Welling}{Kipf and Welling}{2017}]%
        {kipf2017semi}
\bibfield{author}{\bibinfo{person}{Thomas~N. Kipf} {and} \bibinfo{person}{Max
  Welling}.} \bibinfo{year}{2017}\natexlab{}.
\newblock \showarticletitle{Semi-Supervised Classification with Graph
  Convolutional Networks}. In \bibinfo{booktitle}{\emph{International
  Conference on Learning Representations (ICLR)}}.
\newblock


\bibitem[\protect\citeauthoryear{Kulkarni, Praneet, Raghav, and Das}{Kulkarni
  et~al\mbox{.}}{2020}]%
        {kulkarni2020deep}
\bibfield{author}{\bibinfo{person}{Prakhyat~G Kulkarni}, \bibinfo{person}{SY
  Praneet}, \bibinfo{person}{RB Raghav}, {and} \bibinfo{person}{Bhaskarjyoti
  Das}.} \bibinfo{year}{2020}\natexlab{}.
\newblock \showarticletitle{Deep Detection of Anomalies in Static Attributed
  Graph}. In \bibinfo{booktitle}{\emph{International Conference on Machine
  Learning, Image Processing, Network Security and Data Sciences}}. Springer,
  \bibinfo{pages}{627--640}.
\newblock


\bibitem[\protect\citeauthoryear{Kumagai, Iwata, and Fujiwara}{Kumagai
  et~al\mbox{.}}{2020}]%
        {kumagai2020semi}
\bibfield{author}{\bibinfo{person}{Atsutoshi Kumagai},
  \bibinfo{person}{Tomoharu Iwata}, {and} \bibinfo{person}{Yasuhiro Fujiwara}.}
  \bibinfo{year}{2020}\natexlab{}.
\newblock \showarticletitle{Semi-supervised Anomaly Detection on Attributed
  Graphs}.
\newblock \bibinfo{journal}{\emph{arXiv preprint arXiv:2002.12011}}
  (\bibinfo{year}{2020}).
\newblock


\bibitem[\protect\citeauthoryear{Li, Huang, Li, Du, and Zou}{Li
  et~al\mbox{.}}{2019}]%
        {li2019specae}
\bibfield{author}{\bibinfo{person}{Yuening Li}, \bibinfo{person}{Xiao Huang},
  \bibinfo{person}{Jundong Li}, \bibinfo{person}{Mengnan Du}, {and}
  \bibinfo{person}{Na Zou}.} \bibinfo{year}{2019}\natexlab{}.
\newblock \showarticletitle{SpecAE: Spectral AutoEncoder for Anomaly Detection
  in Attributed Networks}. In \bibinfo{booktitle}{\emph{Proceedings of the 28th
  ACM International Conference on Information and Knowledge Management}}.
  \bibinfo{pages}{2233--2236}.
\newblock


\bibitem[\protect\citeauthoryear{Liu, Li, Hu, Fu, Gu, and Xiong}{Liu
  et~al\mbox{.}}{2019}]%
        {liu2019joint}
\bibfield{author}{\bibinfo{person}{Hao Liu}, \bibinfo{person}{Ting Li},
  \bibinfo{person}{Renjun Hu}, \bibinfo{person}{Yanjie Fu},
  \bibinfo{person}{Jingjing Gu}, {and} \bibinfo{person}{Hui Xiong}.}
  \bibinfo{year}{2019}\natexlab{}.
\newblock \showarticletitle{Joint representation learning for multi-modal
  transportation recommendation}. In \bibinfo{booktitle}{\emph{Proceedings of
  the AAAI Conference on Artificial Intelligence}}, Vol.~\bibinfo{volume}{33}.
  \bibinfo{pages}{1036--1043}.
\newblock


\bibitem[\protect\citeauthoryear{Liu, Zheng, Chawla, Yuan, and Xing}{Liu
  et~al\mbox{.}}{2011}]%
        {liu2011discovering}
\bibfield{author}{\bibinfo{person}{Wei Liu}, \bibinfo{person}{Yu Zheng},
  \bibinfo{person}{Sanjay Chawla}, \bibinfo{person}{Jing Yuan}, {and}
  \bibinfo{person}{Xie Xing}.} \bibinfo{year}{2011}\natexlab{}.
\newblock \showarticletitle{Discovering spatio-temporal causal interactions in
  traffic data streams}. In \bibinfo{booktitle}{\emph{Proceedings of the 17th
  ACM SIGKDD International Conference on Knowledge Discovery and Data Mining}}.
  ACM, \bibinfo{pages}{1010--1018}.
\newblock


\bibitem[\protect\citeauthoryear{Malhotra, Ramakrishnan, Anand, Vig, Agarwal,
  and Shroff}{Malhotra et~al\mbox{.}}{2016}]%
        {malhotra2016lstm}
\bibfield{author}{\bibinfo{person}{Pankaj Malhotra}, \bibinfo{person}{Anusha
  Ramakrishnan}, \bibinfo{person}{Gaurangi Anand}, \bibinfo{person}{Lovekesh
  Vig}, \bibinfo{person}{Puneet Agarwal}, {and} \bibinfo{person}{Gautam
  Shroff}.} \bibinfo{year}{2016}\natexlab{}.
\newblock \showarticletitle{LSTM-based encoder-decoder for multi-sensor anomaly
  detection}.
\newblock \bibinfo{journal}{\emph{arXiv preprint arXiv:1607.00148}}
  (\bibinfo{year}{2016}).
\newblock


\bibitem[\protect\citeauthoryear{Mohamed, Qian, Elhoseiny, and Claudel}{Mohamed
  et~al\mbox{.}}{2020}]%
        {mohamed2020social}
\bibfield{author}{\bibinfo{person}{Abduallah Mohamed}, \bibinfo{person}{Kun
  Qian}, \bibinfo{person}{Mohamed Elhoseiny}, {and} \bibinfo{person}{Christian
  Claudel}.} \bibinfo{year}{2020}\natexlab{}.
\newblock \showarticletitle{Social-STGCNN: A Social Spatio-Temporal Graph
  Convolutional Neural Network for Human Trajectory Prediction}. In
  \bibinfo{booktitle}{\emph{Proceedings of the IEEE/CVF Conference on Computer
  Vision and Pattern Recognition}}. \bibinfo{pages}{14424--14432}.
\newblock


\bibitem[\protect\citeauthoryear{Nashville}{Nashville}{2019}]%
        {NFL2019}
\bibfield{author}{\bibinfo{person}{Nashville}.}
  \bibinfo{year}{2019}\natexlab{}.
\newblock \bibinfo{title}{{2019 NFL Draft Event Schedule}}.
\newblock
\newblock
\urldef\tempurl%
\url{https://www.visitmusiccity.com/nfldraft/nfl-draft-event-schedule}
\showURL{%
\tempurl}


\bibitem[\protect\citeauthoryear{Sakurada and Yairi}{Sakurada and
  Yairi}{2014}]%
        {sakurada2014anomaly}
\bibfield{author}{\bibinfo{person}{Mayu Sakurada} {and}
  \bibinfo{person}{Takehisa Yairi}.} \bibinfo{year}{2014}\natexlab{}.
\newblock \showarticletitle{Anomaly detection using autoencoders with nonlinear
  dimensionality reduction}. In \bibinfo{booktitle}{\emph{Proceedings of the
  MLSDA 2014 2nd Workshop on Machine Learning for Sensory Data Analysis}}.
  \bibinfo{pages}{4--11}.
\newblock


\bibitem[\protect\citeauthoryear{Schlichtkrull, Kipf, Bloem, Van Den~Berg,
  Titov, and Welling}{Schlichtkrull et~al\mbox{.}}{2018}]%
        {schlichtkrull2018modeling}
\bibfield{author}{\bibinfo{person}{Michael Schlichtkrull},
  \bibinfo{person}{Thomas~N Kipf}, \bibinfo{person}{Peter Bloem},
  \bibinfo{person}{Rianne Van Den~Berg}, \bibinfo{person}{Ivan Titov}, {and}
  \bibinfo{person}{Max Welling}.} \bibinfo{year}{2018}\natexlab{}.
\newblock \showarticletitle{Modeling relational data with graph convolutional
  networks}. In \bibinfo{booktitle}{\emph{European Semantic Web Conference}}.
  Springer, \bibinfo{pages}{593--607}.
\newblock


\bibitem[\protect\citeauthoryear{Srivastava, Hinton, Krizhevsky, Sutskever, and
  Salakhutdinov}{Srivastava et~al\mbox{.}}{2014}]%
        {srivastava2014dropout}
\bibfield{author}{\bibinfo{person}{Nitish Srivastava},
  \bibinfo{person}{Geoffrey Hinton}, \bibinfo{person}{Alex Krizhevsky},
  \bibinfo{person}{Ilya Sutskever}, {and} \bibinfo{person}{Ruslan
  Salakhutdinov}.} \bibinfo{year}{2014}\natexlab{}.
\newblock \showarticletitle{Dropout: a simple way to prevent neural networks
  from overfitting}.
\newblock \bibinfo{journal}{\emph{The journal of machine learning research}}
  \bibinfo{volume}{15}, \bibinfo{number}{1} (\bibinfo{year}{2014}),
  \bibinfo{pages}{1929--1958}.
\newblock


\bibitem[\protect\citeauthoryear{Su, Zhao, Niu, Liu, Sun, and Pei}{Su
  et~al\mbox{.}}{2019}]%
        {su2019robust}
\bibfield{author}{\bibinfo{person}{Ya Su}, \bibinfo{person}{Youjian Zhao},
  \bibinfo{person}{Chenhao Niu}, \bibinfo{person}{Rong Liu},
  \bibinfo{person}{Wei Sun}, {and} \bibinfo{person}{Dan Pei}.}
  \bibinfo{year}{2019}\natexlab{}.
\newblock \showarticletitle{Robust anomaly detection for multivariate time
  series through stochastic recurrent neural network}. In
  \bibinfo{booktitle}{\emph{Proceedings of the 25th ACM SIGKDD International
  Conference on Knowledge Discovery \& Data Mining}}.
  \bibinfo{pages}{2828--2837}.
\newblock


\bibitem[\protect\citeauthoryear{TDOT}{TDOT}{2019}]%
        {TDOT2019}
\bibfield{author}{\bibinfo{person}{TDOT}.} \bibinfo{year}{2019}\natexlab{}.
\newblock \bibinfo{title}{{TDOT Weekly Construction Report for Middle
  Tennessee}}.
\newblock
\newblock
\urldef\tempurl%
\url{https://preprod.tn.gov/tdot/news/2019/4/17/tdot-weekly-construction-report-for-middle-tennessee--april-18-24--2019.html}
\showURL{%
\tempurl}


\bibitem[\protect\citeauthoryear{Tian, Gao, Cui, Chen, and Liu}{Tian
  et~al\mbox{.}}{2014}]%
        {tian2014learning}
\bibfield{author}{\bibinfo{person}{Fei Tian}, \bibinfo{person}{Bin Gao},
  \bibinfo{person}{Qing Cui}, \bibinfo{person}{Enhong Chen}, {and}
  \bibinfo{person}{Tie-Yan Liu}.} \bibinfo{year}{2014}\natexlab{}.
\newblock \showarticletitle{Learning deep representations for graph
  clustering.}. In \bibinfo{booktitle}{\emph{AAAI}}, Vol.~\bibinfo{volume}{14}.
  \bibinfo{pages}{1293--1299}.
\newblock


\bibitem[\protect\citeauthoryear{TLC}{TLC}{2020}]%
        {NYC}
\bibfield{author}{\bibinfo{person}{NYC TLC}.} \bibinfo{year}{2020}\natexlab{}.
\newblock \bibinfo{title}{TLC Trip Record Data}.
\newblock
  \bibinfo{howpublished}{\url{https://www1.nyc.gov/site/tlc/index.page}}.
\newblock


\bibitem[\protect\citeauthoryear{Uber}{Uber}{2020}]%
        {UM}
\bibfield{author}{\bibinfo{person}{Uber}.} \bibinfo{year}{2020}\natexlab{}.
\newblock \bibinfo{title}{Uber Movement}.
\newblock \bibinfo{howpublished}{\url{https://movement.uber.com/}}.
\newblock


\bibitem[\protect\citeauthoryear{Veli{\v{c}}kovi{\'c}, Cucurull, Casanova,
  Romero, Lio, and Bengio}{Veli{\v{c}}kovi{\'c} et~al\mbox{.}}{2017}]%
        {velivckovic2017graph}
\bibfield{author}{\bibinfo{person}{Petar Veli{\v{c}}kovi{\'c}},
  \bibinfo{person}{Guillem Cucurull}, \bibinfo{person}{Arantxa Casanova},
  \bibinfo{person}{Adriana Romero}, \bibinfo{person}{Pietro Lio}, {and}
  \bibinfo{person}{Yoshua Bengio}.} \bibinfo{year}{2017}\natexlab{}.
\newblock \showarticletitle{Graph attention networks}.
\newblock \bibinfo{journal}{\emph{arXiv preprint arXiv:1710.10903}}
  (\bibinfo{year}{2017}).
\newblock


\bibitem[\protect\citeauthoryear{Wang, Pan, Long, Zhu, and Jiang}{Wang
  et~al\mbox{.}}{2017}]%
        {wang2017mgae}
\bibfield{author}{\bibinfo{person}{Chun Wang}, \bibinfo{person}{Shirui Pan},
  \bibinfo{person}{Guodong Long}, \bibinfo{person}{Xingquan Zhu}, {and}
  \bibinfo{person}{Jing Jiang}.} \bibinfo{year}{2017}\natexlab{}.
\newblock \showarticletitle{Mgae: Marginalized graph autoencoder for graph
  clustering}. In \bibinfo{booktitle}{\emph{Proceedings of the 2017 ACM on
  Conference on Information and Knowledge Management}}.
  \bibinfo{pages}{889--898}.
\newblock


\bibitem[\protect\citeauthoryear{Wang, Yin, Chen, Wo, Xu, and Zheng}{Wang
  et~al\mbox{.}}{2019b}]%
        {wang2019origin}
\bibfield{author}{\bibinfo{person}{Yuandong Wang}, \bibinfo{person}{Hongzhi
  Yin}, \bibinfo{person}{Hongxu Chen}, \bibinfo{person}{Tianyu Wo},
  \bibinfo{person}{Jie Xu}, {and} \bibinfo{person}{Kai Zheng}.}
  \bibinfo{year}{2019}\natexlab{b}.
\newblock \showarticletitle{Origin-destination matrix prediction via graph
  convolution: a new perspective of passenger demand modeling}. In
  \bibinfo{booktitle}{\emph{Proceedings of the 25th ACM SIGKDD International
  Conference on Knowledge Discovery \& Data Mining}}.
  \bibinfo{pages}{1227--1235}.
\newblock


\bibitem[\protect\citeauthoryear{Wang, Ren, He, Zhang, and Hu}{Wang
  et~al\mbox{.}}{2019a}]%
        {wang2019robust}
\bibfield{author}{\bibinfo{person}{Zihan Wang}, \bibinfo{person}{Zhaochun Ren},
  \bibinfo{person}{Chunyu He}, \bibinfo{person}{Peng Zhang}, {and}
  \bibinfo{person}{Yue Hu}.} \bibinfo{year}{2019}\natexlab{a}.
\newblock \showarticletitle{Robust Embedding with Multi-Level Structures for
  Link Prediction.}. In \bibinfo{booktitle}{\emph{IJCAI}}.
  \bibinfo{pages}{5240--5246}.
\newblock


\bibitem[\protect\citeauthoryear{Wu, Pan, Chen, Long, Zhang, and Philip}{Wu
  et~al\mbox{.}}{2020}]%
        {wu2020comprehensive}
\bibfield{author}{\bibinfo{person}{Zonghan Wu}, \bibinfo{person}{Shirui Pan},
  \bibinfo{person}{Fengwen Chen}, \bibinfo{person}{Guodong Long},
  \bibinfo{person}{Chengqi Zhang}, {and} \bibinfo{person}{S~Yu Philip}.}
  \bibinfo{year}{2020}\natexlab{}.
\newblock \showarticletitle{A comprehensive survey on graph neural networks}.
\newblock \bibinfo{journal}{\emph{IEEE Transactions on Neural Networks and
  Learning Systems}} (\bibinfo{year}{2020}).
\newblock


\bibitem[\protect\citeauthoryear{Xu, Chen, Zhao, Li, Bu, Li, Liu, Zhao, Pei,
  Feng, et~al\mbox{.}}{Xu et~al\mbox{.}}{2018}]%
        {xu2018unsupervised}
\bibfield{author}{\bibinfo{person}{Haowen Xu}, \bibinfo{person}{Wenxiao Chen},
  \bibinfo{person}{Nengwen Zhao}, \bibinfo{person}{Zeyan Li},
  \bibinfo{person}{Jiahao Bu}, \bibinfo{person}{Zhihan Li},
  \bibinfo{person}{Ying Liu}, \bibinfo{person}{Youjian Zhao},
  \bibinfo{person}{Dan Pei}, \bibinfo{person}{Yang Feng}, {et~al\mbox{.}}}
  \bibinfo{year}{2018}\natexlab{}.
\newblock \showarticletitle{Unsupervised anomaly detection via variational
  auto-encoder for seasonal {KPI}s in web applications}. In
  \bibinfo{booktitle}{\emph{Proceedings of the 2018 World Wide Web
  Conference}}. \bibinfo{pages}{187--196}.
\newblock


\bibitem[\protect\citeauthoryear{Yang, Yih, He, Gao, and Deng}{Yang
  et~al\mbox{.}}{2014b}]%
        {yang2014embedding}
\bibfield{author}{\bibinfo{person}{Bishan Yang}, \bibinfo{person}{Wen-tau Yih},
  \bibinfo{person}{Xiaodong He}, \bibinfo{person}{Jianfeng Gao}, {and}
  \bibinfo{person}{Li Deng}.} \bibinfo{year}{2014}\natexlab{b}.
\newblock \showarticletitle{Embedding entities and relations for learning and
  inference in knowledge bases}.
\newblock \bibinfo{journal}{\emph{arXiv preprint arXiv:1412.6575}}
  (\bibinfo{year}{2014}).
\newblock


\bibitem[\protect\citeauthoryear{Yang, Kalpakis, and Biem}{Yang
  et~al\mbox{.}}{2014a}]%
        {yang2014detecting}
\bibfield{author}{\bibinfo{person}{Shiming Yang}, \bibinfo{person}{Konstantinos
  Kalpakis}, {and} \bibinfo{person}{Alain Biem}.}
  \bibinfo{year}{2014}\natexlab{a}.
\newblock \showarticletitle{Detecting road traffic events by coupling multiple
  timeseries with a nonparametric {B}ayesian method}.
\newblock \bibinfo{journal}{\emph{IEEE Transactions on Intelligent
  Transportation Systems}} \bibinfo{volume}{15}, \bibinfo{number}{5}
  (\bibinfo{year}{2014}), \bibinfo{pages}{1936--1946}.
\newblock


\bibitem[\protect\citeauthoryear{Yu, Yin, and Zhu}{Yu et~al\mbox{.}}{2017}]%
        {yu2017spatio}
\bibfield{author}{\bibinfo{person}{Bing Yu}, \bibinfo{person}{Haoteng Yin},
  {and} \bibinfo{person}{Zhanxing Zhu}.} \bibinfo{year}{2017}\natexlab{}.
\newblock \showarticletitle{Spatio-temporal graph convolutional networks: A
  deep learning framework for traffic forecasting}.
\newblock \bibinfo{journal}{\emph{Proceedings of the 27th International Joint
  Conference on Artificial Intelligence}} (\bibinfo{year}{2017}).
\newblock


\bibitem[\protect\citeauthoryear{Yu, Cheng, Aggarwal, Zhang, Chen, and Wang}{Yu
  et~al\mbox{.}}{2018}]%
        {yu2018netwalk}
\bibfield{author}{\bibinfo{person}{Wenchao Yu}, \bibinfo{person}{Wei Cheng},
  \bibinfo{person}{Charu~C Aggarwal}, \bibinfo{person}{Kai Zhang},
  \bibinfo{person}{Haifeng Chen}, {and} \bibinfo{person}{Wei Wang}.}
  \bibinfo{year}{2018}\natexlab{}.
\newblock \showarticletitle{Netwalk: A flexible deep embedding approach for
  anomaly detection in dynamic networks}. In
  \bibinfo{booktitle}{\emph{Proceedings of the 24th ACM SIGKDD International
  Conference on Knowledge Discovery \& Data Mining}}.
  \bibinfo{pages}{2672--2681}.
\newblock


\bibitem[\protect\citeauthoryear{Zhai, Cheng, Lu, and Zhang}{Zhai
  et~al\mbox{.}}{2016}]%
        {zhai2016deep}
\bibfield{author}{\bibinfo{person}{Shuangfei Zhai}, \bibinfo{person}{Yu Cheng},
  \bibinfo{person}{Weining Lu}, {and} \bibinfo{person}{Zhongfei Zhang}.}
  \bibinfo{year}{2016}\natexlab{}.
\newblock \showarticletitle{Deep structured energy based models for anomaly
  detection}. In \bibinfo{booktitle}{\emph{Proceedings of the 33rd
  International Conference on International Conference on Machine
  Learning-Volume 48}}. \bibinfo{pages}{1100--1109}.
\newblock


\bibitem[\protect\citeauthoryear{Zheng, Li, Li, Li, and Gao}{Zheng
  et~al\mbox{.}}{2019}]%
        {zheng2019addgraph}
\bibfield{author}{\bibinfo{person}{Li Zheng}, \bibinfo{person}{Zhenpeng Li},
  \bibinfo{person}{Jian Li}, \bibinfo{person}{Zhao Li}, {and}
  \bibinfo{person}{Jun Gao}.} \bibinfo{year}{2019}\natexlab{}.
\newblock \showarticletitle{AddGraph: Anomaly Detection in Dynamic Graph Using
  Attention-based Temporal GCN.}. In \bibinfo{booktitle}{\emph{IJCAI}}.
  \bibinfo{pages}{4419--4425}.
\newblock


\bibitem[\protect\citeauthoryear{Zhou and Paffenroth}{Zhou and
  Paffenroth}{2017}]%
        {zhou2017anomaly}
\bibfield{author}{\bibinfo{person}{Chong Zhou} {and} \bibinfo{person}{Randy~C
  Paffenroth}.} \bibinfo{year}{2017}\natexlab{}.
\newblock \showarticletitle{Anomaly detection with robust deep autoencoders}.
  In \bibinfo{booktitle}{\emph{Proceedings of the 23rd ACM SIGKDD International
  Conference on Knowledge Discovery and Data Mining}}.
  \bibinfo{pages}{665--674}.
\newblock


\bibitem[\protect\citeauthoryear{Zhou, Yang, Ren, Wu, and Zhuang}{Zhou
  et~al\mbox{.}}{2018}]%
        {zhou2018dynamic}
\bibfield{author}{\bibinfo{person}{Le-kui Zhou}, \bibinfo{person}{Yang Yang},
  \bibinfo{person}{Xiang Ren}, \bibinfo{person}{Fei Wu}, {and}
  \bibinfo{person}{Yueting Zhuang}.} \bibinfo{year}{2018}\natexlab{}.
\newblock \showarticletitle{Dynamic Network Embedding by Modeling Triadic
  Closure Process.}. In \bibinfo{booktitle}{\emph{AAAI}}.
  \bibinfo{pages}{571--578}.
\newblock


\bibitem[\protect\citeauthoryear{Zhu, Ma, and Liu}{Zhu et~al\mbox{.}}{2020}]%
        {zhu2020deepad}
\bibfield{author}{\bibinfo{person}{Dali Zhu}, \bibinfo{person}{Yuchen Ma},
  {and} \bibinfo{person}{Yinlong Liu}.} \bibinfo{year}{2020}\natexlab{}.
\newblock \showarticletitle{DeepAD: A Joint Embedding Approach for Anomaly
  Detection on Attributed Networks}. In \bibinfo{booktitle}{\emph{International
  Conference on Computational Science}}. Springer, \bibinfo{pages}{294--307}.
\newblock


\bibitem[\protect\citeauthoryear{Zong, Song, Min, Cheng, Lumezanu, Cho, and
  Chen}{Zong et~al\mbox{.}}{2018}]%
        {zong2018deep}
\bibfield{author}{\bibinfo{person}{Bo Zong}, \bibinfo{person}{Qi Song},
  \bibinfo{person}{Martin~Renqiang Min}, \bibinfo{person}{Wei Cheng},
  \bibinfo{person}{Cristian Lumezanu}, \bibinfo{person}{Daeki Cho}, {and}
  \bibinfo{person}{Haifeng Chen}.} \bibinfo{year}{2018}\natexlab{}.
\newblock \showarticletitle{Deep autoencoding gaussian mixture model for
  unsupervised anomaly detection}. In \bibinfo{booktitle}{\emph{International
  Conference on Learning Representations}}.
\newblock


\end{thebibliography}

\clearpage
\appendix

\section{Supplementary Material}

\subsection{Experiment data preparation and simulation}
In this section, we detail the data reprocessing and simulation steps. We also show the data distribution of travel times for the three datasets used in the paper.

As described in the main text, the dataset is split into training and testing sets. The training set is assumed to have no major events (anomalies). To compare methods in a controlled setting, we create a test set with synthetic labeled anomalies which are injected into cleaned traffic conditions. We synthetically generate the test set from real data due to the unavailability of city scale traffic datasets pre-labeled with anomalies. 

To generate cleaned traffic conditions in the test set which are later polluted with anomalies, we assume the travel time of each OD-pair in each hour of day in each day of week follows a Gaussian distribution. We calculate the mean and variance of the OD pair travel time of corresponding hours from the raw test data from the first quarter of 2019. We then resample the travel times based on the calculated mean and variance to arrive at a complete clean test dataset. Next we inject spatial and temporal anomalies respectively as described in the main text to arrive at the labeled test set with anomalies.

In the Uber Movement dataset corresponding to Nashville, TN, the region is divided into 2219 zones. However many of the OD pairs never record travel times. To learn as much useful information as possible and ensure learning efficiency, we select the top 50 most connected zones, so that the selected zone has one or more trip to at least one third of the total 2219 zones. This results in 41\% missing data rate in the training set, and a 30\% missing data rate in the test set. The minimum and maximum latitude and longitude extents of each zone are used as the node features, which are scaled to 0-1 before feeding into \GNN.

\begin{figure}
\centering

\begin{subfigure}[t]{\columnwidth}
    \includegraphics[width=\linewidth]{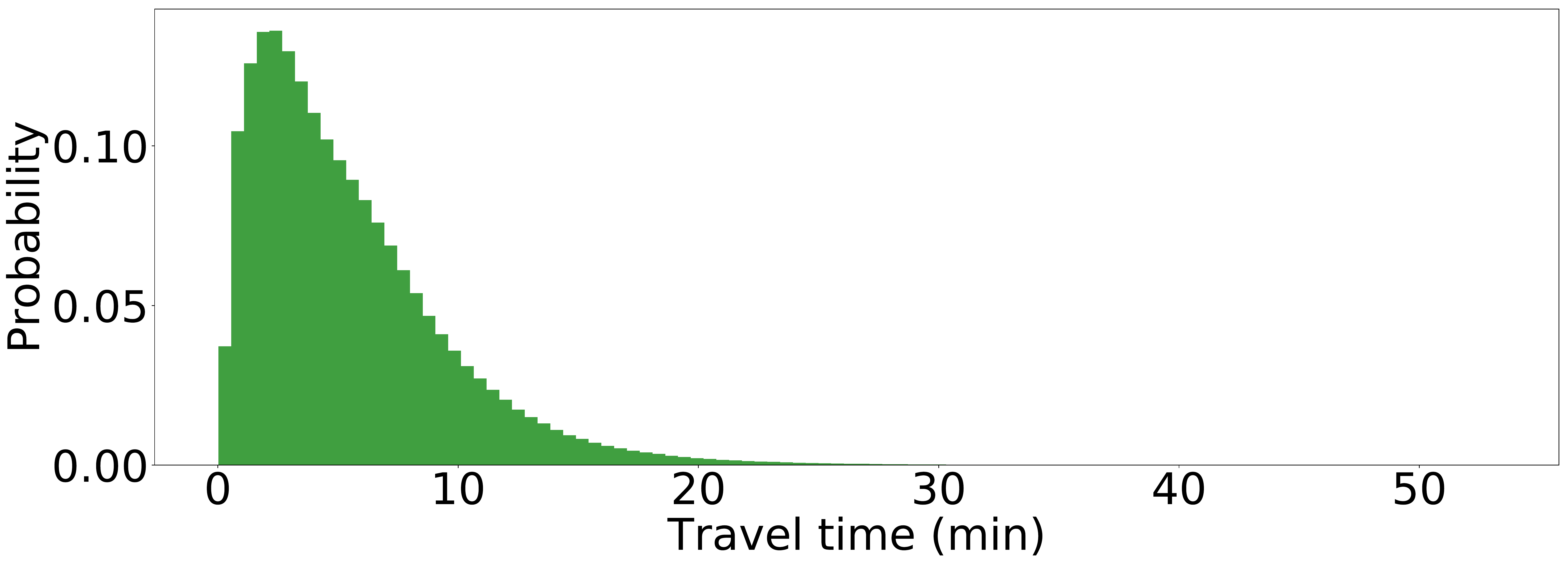}
    \caption{Travel time PDF of Uber Movement data}
\end{subfigure}
\quad
\begin{subfigure}[t]{\columnwidth}
    \includegraphics[width=\linewidth]{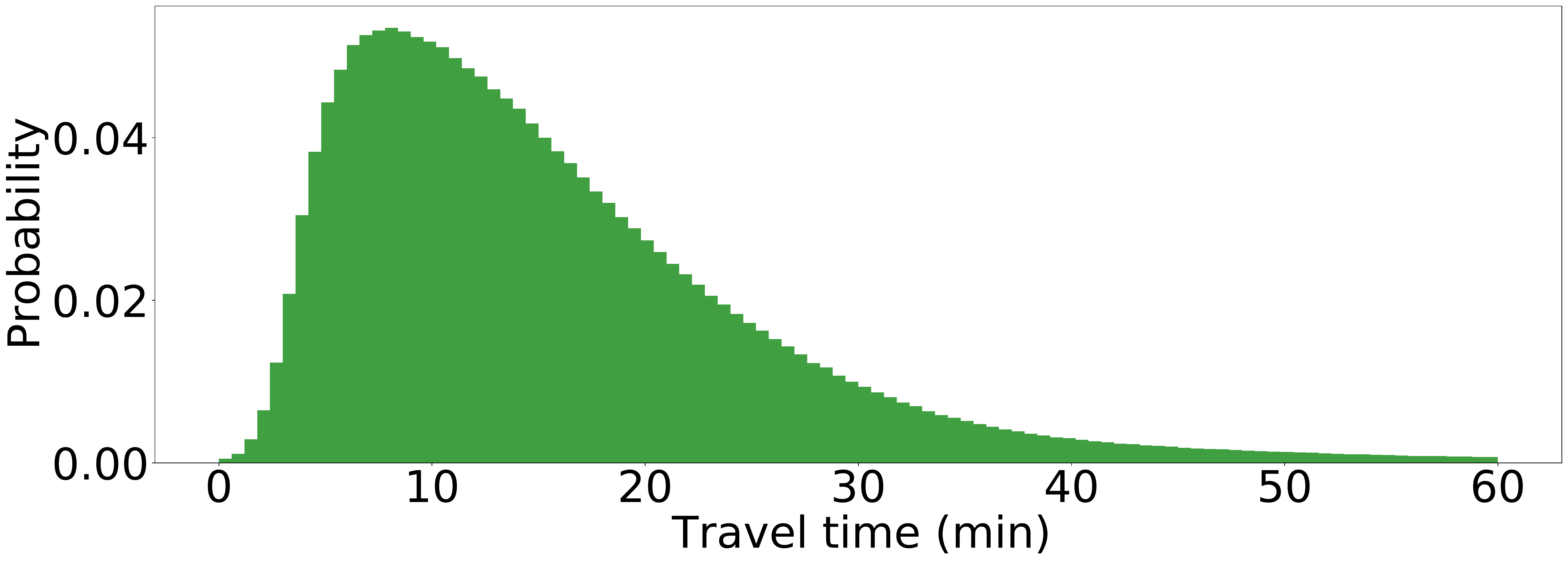}
    \caption{Travel time PDF of NYC taxi data}
\end{subfigure}
\quad
\begin{subfigure}[t]{\columnwidth}
    \includegraphics[width=\linewidth]{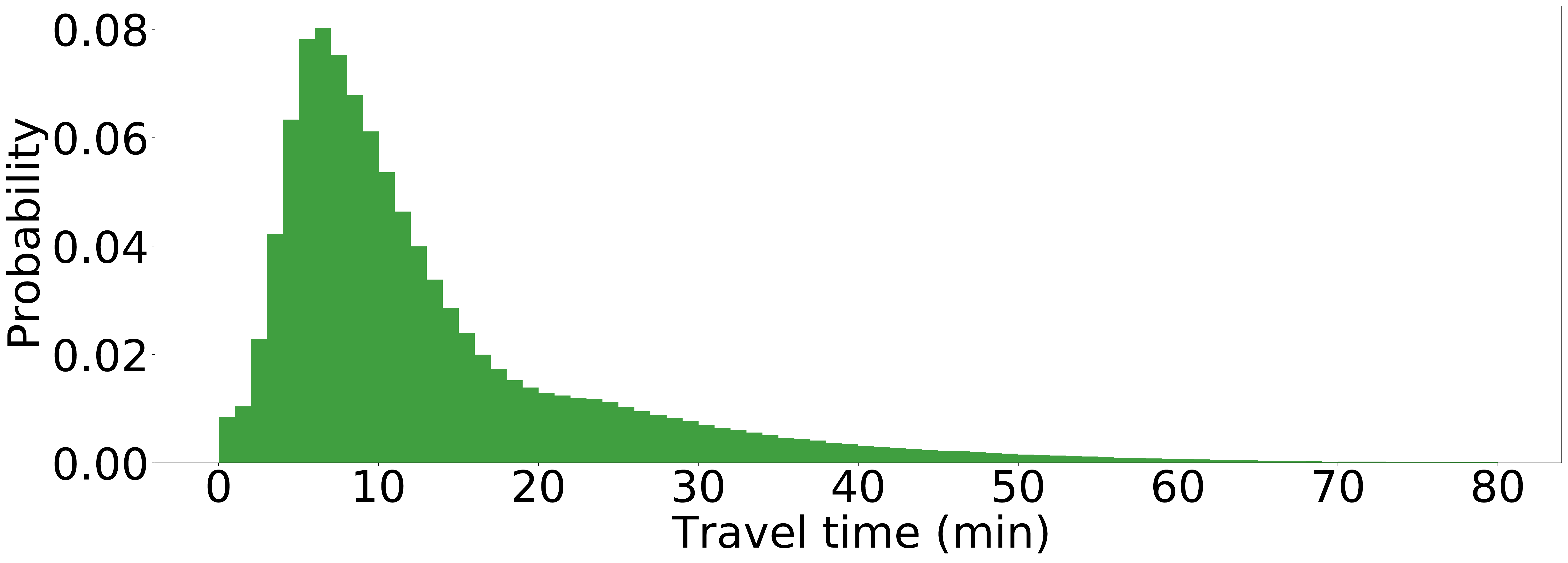}
    \caption{Travel time PDF of Chicago taxi data}
\end{subfigure}
\quad
\caption{Travel time distribution for three datasets.}
\label{fig:distr}
\end{figure}

The data distribution of travel times for the three datasets used in the paper are shown in Fig.~\ref{fig:distr}. We can see that Uber Movement dataset contains trips of shorter duration than NYC or Chicago taxi data. Meanwhile, Uber data and Chicago taxi data has more concentrated travel time distribution than NYC taxi data.

\subsection{Baseline methods}
In this section, we give a brief introduction to each method, and document the model configurations.
\begin{itemize}
    \item RTC~\cite{hu2019robust}. Robust tensor recovery and completion for group anomaly detection (RTC) is a low-rank tensor decomposition based method that exploits spatio-temporal correlation. This a non-deep learning baseline method.
    
    RTC has one hyper parameter $\lambda$, which controls the tradeoff between the rank of the normal tensor and the size and magnitude of the outlier tensor. We set $\lambda = 0.278$ for Uber data, set $\lambda = 0.390$ for NYC data, and set $\lambda = 0.227$ for Chicago data. The value is chosen empirically following the settings described in the article.
    
    
    \item AE~\cite{hawkins2002outlier}. Autoencoder (AE) is a widely-used neural network model for anomaly detection. It uses a fully connected neural network to build an encoder and a decoder to compress and reconstruct the data. It uses the reconstruction error as the anomaly score.
    
    We set the hidden size (i.e., the dimension of the low dimensional encoded state) of AE as the same setting used in \GNN. Namely, the hidden size is 150 for Uber data, 50 for NYC, and 25 for Chicago data. The learning rate is 0.0005, and the number of epochs is 30. The values were determined to maximize performance using grid search. The sequential length is set as 1, i.e., each input sample contains the traffic conditions of one time step. All other parameters are set as the default values, with random seed set at 0.
    
    \item EncDec-AD~\cite{malhotra2016lstm}. EncDec-AD is an LSTM-based encoder-decoder model for time series anomaly detection. EncDec-AD learns the normal pattern of a multi-dimensional time series and detects anomalous times that deviate from the normal pattern via the reconstruction error.
    
    We set the hidden size (i.e., the dimension of the low dimensional encoded state) of AE as the same setting used in \GNN. Namely, the hidden size is 150 for Uber data, 50 for NYC, and 25 for Chicago data. The learning rate and number of epochs are 0.0001 and 120 for Uber data, 0.0005 and 30 for NYC, and 0.001 and 70 for Chicago data. EncDec-AD takes time series windows as input, and we set the sequential length at 24 for Uber and Chicago data, and 72 for NYC data, which are also determined via grid search. All other parameters are set as the default values, with random seed set at 0.
    
    \item DAGMM~\cite{zong2018deep}. \textit{Deep autoencoding Gaussian mixture model} (DAGMM) conducts unsupervised anomaly detection by finding low-dimensional data representations via a deep autoencoder, and modeling the data distribution in the low-dimensional space via a Gaussian mixture model.
    
    Instead of setting the hidden size of DAGMM at 150 like the \GNN,  a grid search revealed the best performance is achieved at a hidden size of 10 for Uber data, 50 for NYC data, and 5 for Chicago data. While the other methods (\GNN, AE, EncDec-AD) are reconstruction-based, DAGMM relies on clustering in a low-dimensional space. An encoding dimension too low might not capture the original information, while high dimensional clustering can encounter a curse of dimensionality.  The learning rate and number of epochs are 0.0001 and 20 for Uber data, 0.0005 and 50 for NYC data, and 0.0001 and 20 for Chicago data, which yields the best result after parameter grid search. The sequential length is set as 1, i.e., each input sample contains the traffic conditions of one time step. The rest parameters are set at the default values, with the random seed at 0.
    
    \item REBM~\cite{zhai2016deep}. REBM is the RNN-based version of \textit{deep structured energy based models} (EBM) for time series anomaly detection. REBM models the data distribution with neural networks, and detects the anomaly by the energy (negative log probability) and reconstruction error.
    
     We set the hidden size (i.e., the dimension of the low dimensional encoded state) of AE as the same setting used in \GNN. Namely, the hidden size is 150 for Uber data, 50 for NYC, and 25 for Chicago data. The learning rate and number of epochs are 0.001 and 100 for Uber data, 0.0001 and 100 for NYC, and 0.001 and 50 for Chicago data. All other parameters are set as the default values, with random seed set at 0.
    
    
    \item GCN~\cite{kipf2017semi}. As a baseline formulation of Graph neural network, GCN updates node features by averaging over the features of neighbor nodes. We use the final node feature matrix of all nodes as the encoding for the traffic graph, and use the same decoder as our Con-GAE model to decode the edge weights from the node encoding. The reconstruction error across all edges is used as anomaly score. 
    
    For GCN we use the same structure as \GNN model. Namely, we use two layers of GCN, the node embeddings are set at 300 and 150 for Uber data; 150 and 50 for NYC data, and 300 and 25 for Chicago data. The dropout rate is set at 0.2. The learning rate is set at 0.0001 for Uber data, 0.001 for NYC data, and 0.01 for Chicago data.  
    
    \item GraphSAGE~\cite{hamilton2017inductive} improves from GCN by explicitly concatenating the feature of the node itself after aggregating the neighboring node features. Its encoding-decoding structure is the same as our GCN comparison method. It is akin to our Con-GAE model without the aggregating layer from node embedding to graph embedding and without the time embedding.
    
    For GraphSAGE we use the same structure as \GNN model. Namely, we use two layers of GCN, the node embeddings for the two layers are set at 300 and 150 for Uber data; 150 and 50 for NYC data, and 300 and 25 for Chicago data. The dropout rate is set at 0.2. The learning rate is set at 0.0001 for Uber data, 0.0005 for NYC data, and 0.05 for Chicago data.
    
\end{itemize}

For the methods that do not directly accommodate missing data, namely HA, AE, DAGMM, EncDec-AD and REBM, we impute the missing data via temporal interpolation (i.e., via linear interpolation when values are present to interpolate, or via a mean value when interpolation cannot be performed).

\subsection{Reproducibility}
In this section, we specify the computing hardware and software used for experiments, as well as the data availability.

The GPU used for the experiments is a GTX 1080 (12GB) operated on Ubuntu 18.04.4 LTS. We use the python package Pytorch Geometric~\cite{Fey/Lenssen/2019} version 1.4.3 to implement the GCN.


The Uber Movement time series data is from Nashville, TN, which is not yet publicly available on the Uber Movement portal~\cite{UM}. Uber Movement is planning to release the data on the portal, timed with the publication of this work. Uber Movement data is available under a Creative Commons, Attribution Non-Commercial license. \footnote{\url{https://movement.uber.com/faqs?lang=en-US}}. 

\end{document}